\documentclass[journal]{IEEEtran}
\usepackage{times}  
\usepackage{helvet} 
\usepackage{courier}  
\usepackage[hyphens]{url}  
\usepackage{graphicx} 
\usepackage{graphicx}  
\usepackage{lipsum}
\usepackage{amsmath}
\usepackage{amssymb}
\usepackage{mathtools}
\usepackage{adjustbox}
\usepackage{gensymb}
\usepackage{indentfirst}
\usepackage{multirow}
\usepackage{etoolbox}
\usepackage{subfigure}
\usepackage{xcolor}
\newcommand{\eat}[1]{}

\usepackage{algorithm}
\usepackage{algorithmic}

\usepackage{multirow}
\usepackage{footnote}

\usepackage{subfigure}
\usepackage{framed}
\usepackage{colortbl} 
\usepackage{arydshln} 
\definecolor{mygray}{gray}{.9}
\definecolor{mypink}{rgb}{.99,.91,.95}
\usepackage{bbding}
\usepackage{color}
\usepackage{booktabs}
\newcommand{\keypoint}[1]{\vspace{0.1cm}\noindent\textbf{#1}}
\usepackage{bbm}
\usepackage{color}

\usepackage[colorlinks=true,linkcolor=red,citecolor=red,urlcolor=magenta,]{hyperref}
\usepackage[doipre={doi:~}]{uri}

\begin{document}
 \title{From Global to Local: \\
 Multi-scale Out-of-distribution Detection}
	\author{Ji Zhang,  
		Lianli Gao,~\IEEEmembership{Member, IEEE},
        Bingguang Hao,
        Hao Huang, \\
        Jingkuan Song,~\IEEEmembership{Senior Member, IEEE} and
		Hengtao Shen,~\IEEEmembership{Fellow, IEEE}
		\thanks{Ji Zhang (jizhang.jim@gmail.com), Lianli Gao, Jingkuan Song and Hengtao Shen are with the School of Computer Science and Engineering, Bingguang Hao is with the Yingcai Honors College, University of Electronic Science and Technology of China. Hao Huang is with Kuaishou Technology.}
		}
	\markboth{}%
	{Shell \MakeLowercase{\textift{et al.}}: Local Propagation for Out-of-distribution Detection}
	\maketitle

	\begin{abstract}
    Out-of-distribution (OOD) detection aims to detect “unknown” data whose labels have not been seen during the in-distribution (ID) training process. Recent progress in representation learning gives rise to distance-based OOD detection that recognizes inputs as ID/OOD according to their relative distances to the training data of ID classes. 
    Previous approaches calculate pairwise distances relying only on global image representations, which can be sub-optimal as the inevitable background clutter and intra-class variation may drive image-level representations from the same ID class far apart in a given representation space. In this work, we overcome this challenge by proposing \underline{Multi-scale OOD DEtection (MODE)}, a first framework leveraging both global visual information and local region details of images to maximally benefit OOD detection. 
    Specifically, we first find that existing models pretrained by off-the-shelf cross-entropy or contrastive losses are incompetent to capture valuable local representations for MODE, due to the scale-discrepancy between the ID training and OOD detection processes. To mitigate this issue and encourage locally discriminative representations in ID training, we propose Attention-based Local PropAgation (\textbf{$\mathtt{ALPA}$}), a trainable objective that exploits a cross-attention mechanism to align and highlight the local regions of the target objects for pairwise examples. 
    During test-time OOD detection, a Cross-Scale Decision (\textbf{$\mathtt{CSD}$}) function is further devised on the most discriminative multi-scale representations to distinguish ID/OOD data more faithfully.  
    We demonstrate the effectiveness and flexibility of MODE on several benchmarks -- on average, MODE outperforms the previous state-of-the-art by up to \underline{\textbf{19.24\%}} in FPR, \underline{\textbf{2.77\%}} in AUROC.
    Code is available at \url{{https://github.com/JimZAI/MODE-OOD}}.
  
	\end{abstract}
	
	\begin{IEEEkeywords}
		Out-of-distribution detection, Outlier Detection, Anomaly Detection, Multi-scale Representations.
	\end{IEEEkeywords}
	
	\IEEEpeerreviewmaketitle

	\section{Introduction}

“No machine is perfect”, modern machine learning (ML) systems are shown to produce overconfident and thus untrustworthy predictions for “unknown” out-of-distribution (OOD) inputs -- whose labels have not been seen during the in-distribution (ID) training procedure \cite{hendrycks2016baseline,lee2018simple,liu2020energy}. 
 This gives rise to a more general and realistic task of OOD detection recently, where the goal is to distinguish whether an incoming example is ID/OOD and allows the ML system to take precautions in deployment \cite{hsu2020generalized,liang2017enhancing,bai2021decaug,salehiunified}.
For instance, in autonomous driving, a safety-critical application, the driving system must hand over the control to drivers when it detects OOD data, e.g., unusual objects, scenes.

\begin{figure}[t]
\setlength{\abovecaptionskip}{0.5cm}  
\setlength{\belowcaptionskip}{-0.5cm} 
    \centering
    \includegraphics[width=0.46\textwidth]{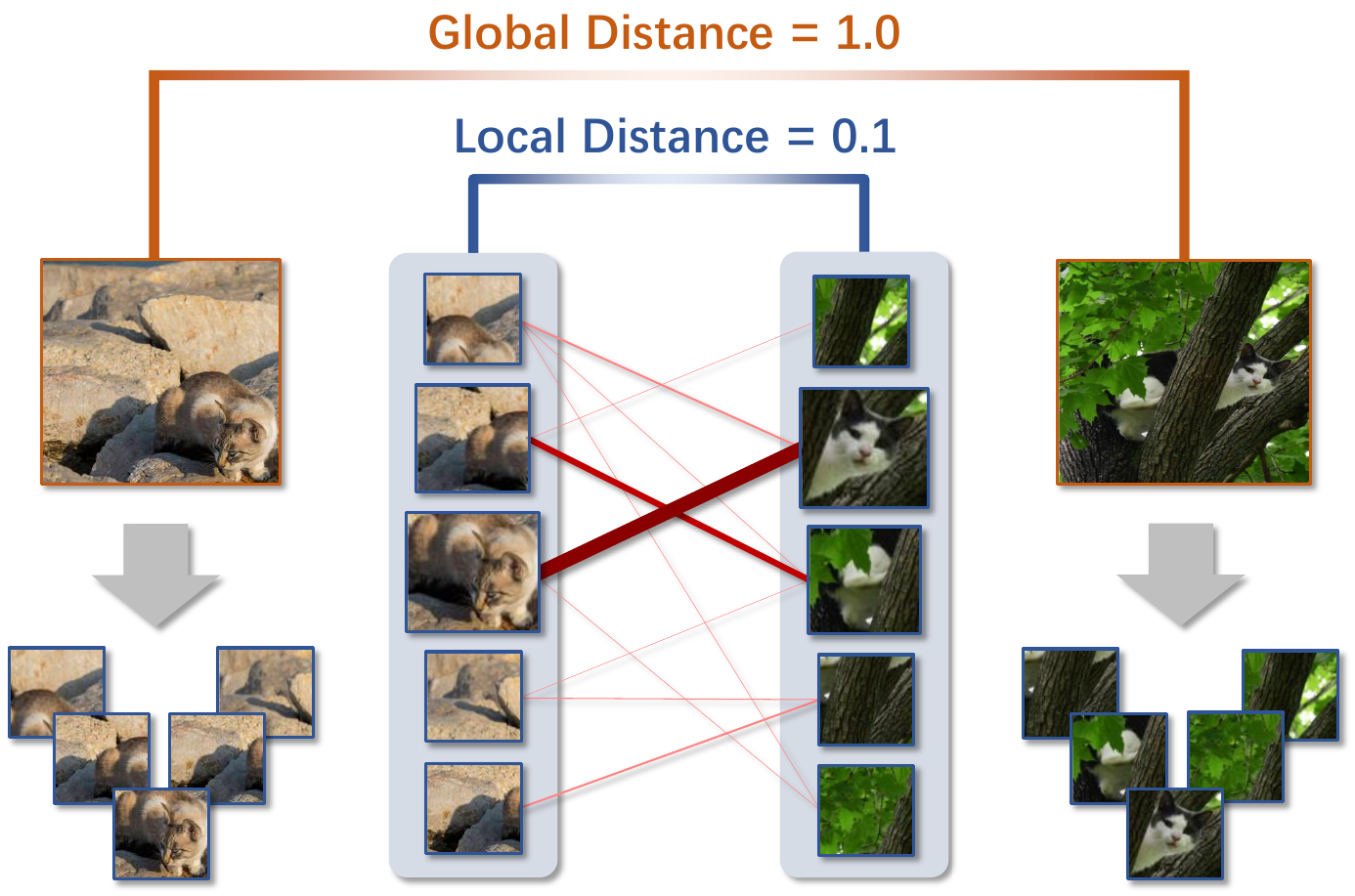}
    \caption{Motivation of exploring local region representations to determine the image relevance between pairwise examples:  the inevitable background clutter and intra-class variation may drive global image representations from the same ID class far apart in a given representation space.
    \textbf{{For the first time, we take advantage of both global visual information and local region details of images to maximally benefit OOD detection.}}
    }
    \label{f1}
\end{figure}

A plethora of OOD detection schemes have been proposed to mitigate the risk of OOD data, while classifying ID classes correctly. 
Many prior works safeguard against OOD examples relying on a softmax scoring mechanism, which is motivated by the simple observation that examples with lower softmax confidence scores are more likely to be OOD \cite{liang2017enhancing,liu2020energy}. 
Nevertheless, well-performed models can produce arbitrarily high softmax confidence for inputs far away from the training data \cite{nguyen2015deep,sun2022out}.
Recent progress in representation learning gives rise to distance-based OOD detection that
represents image data in an appropriate representation space and uses a distance function to decide whether testing examples are ID/OOD according to their relative distances to the training data of ID classes \cite{sehwagssd,tack2020csi,sun2022out}.
Particularly, Sun et al. proposed KNN \cite{sun2022out}, a first work exploring the effectiveness of using a \textit{k}-nearest neighbor search over global image representations (a.k.a. the penultimate layer representations) for OOD detection.
In addition to establishing  state-of-the-art performance on various OOD benchmarks and network structures, several compelling advantages of KNN-based OOD detection, such as \textbf{i}) easy-to-use, \textbf{ii}) model-agnostic, and \textbf{iii}) distribution assumption-free, make it enjoy good practicability and scalability. 

Despite the encouraging advantages, we observe that the inevitable background clutter and intra-class variation may drive the global, image-level representations from the same ID class far apart in a given representation space, as illustrated in Fig. \ref{f1}.
Therefore, it becomes more difficult to effectively distinguish ID-OOD examples, relying only on the \textit{single-scale} global representations.
Furthermore, overwhelming empirical evidence reveals that
exploring richer visual information from \textit{multi-scale} representations is of great importance for understanding discriminative local regions, and semantic categories of the target objects \cite{gao2019res2net,jain1998multiscale}.
However, looking at the literature on OOD detection over the past years, 
the efficiency of exploiting discriminative local representations for achieving better ID-OOD separability has not received any attention so far, not to mention leveraging both global and local representations to maximally benefit OOD detection.
{This limitation begs the following question:}

\begin{framed}
 Can we take advantage of both global visual information and local region details from images to distinguish ID/OOD examples more effectively?
\end{framed}

\begin{figure}[t]
    \centering
    \includegraphics[width=0.49\textwidth]{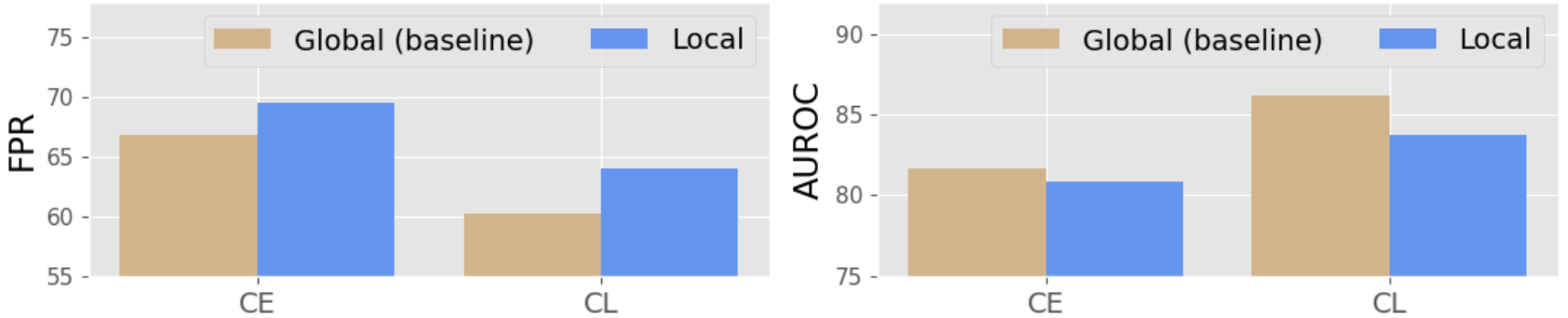}
    \caption{Performance degradation (\textcolor{blue}{blue} bar) caused by the \textbf{scale-discrepancy} between ID training (on  \textit{global} representations) and OOD detection (on  \textit{local} representations) -- {CIFAR-100 (ID) with ResNet-34}. The model is pretrained by cross-entropy (CE) or contrastive learning (CL) loss. The results are the average on the five common OOD datasets shown in Section \ref{exp1}. For  FPR (resp. AUROC), smaller (resp. higher) values indecate better performance.}
    \label{mtv}
\end{figure}

\begin{figure*}[t]
    \centering
    \includegraphics[width=0.98\textwidth]{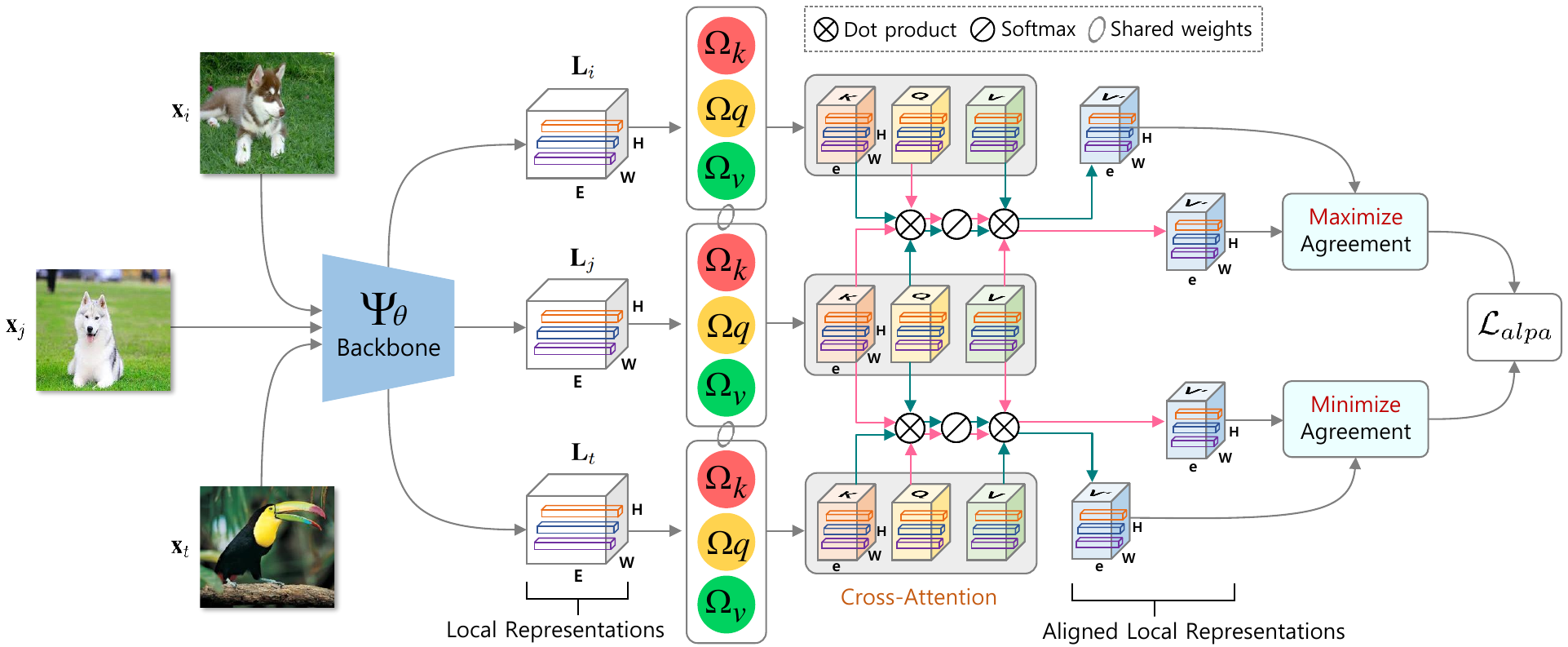}
    \caption{An overview of \textit{training-time} \textbf{Attention-based Local Propagation ($\mathtt{ALPA}$)} that encourages the mining of discriminative local representations for MODE.
    First, the feature backbone $\Psi_{\theta}$ takes each image $\mathbf{\textbf{x}}$ as input to produce the local representations (a.k.a. dense features) $\mathbf{\textbf{L}} = {\Psi_{\theta}}(\mathbf{\textbf{x}}) \in \mathbb{R}^{HW\times E}$. 
    Then, three linear projection heads $\Omega_k$, $\Omega_q$ and $\Omega_v$ transform $\mathbf{\textbf{L}}$ to a lower-dimensional space and obtain the value $\mathbf{\textbf{K}}=\Omega_k(\mathbf{\textbf{L}})$, the query $\mathbf{\textbf{Q}}=\Omega_q(\mathbf{\textbf{L}})$ and the key $\mathbf{\textbf{V}}=\Omega_v(\mathbf{\textbf{L}})$ respectively, where $\mathbf{\textbf{K}}$,$\mathbf{\textbf{Q}}$,$\mathbf{\textbf{V}}\in \mathbb{R}^{HW \times e}$.
    Next, a cross-attention mechanism is applied to align the $e$-dimensional local representations of pairwise examples, so as to highlight the target object regions.
    Finally, the parameters of $\Psi_{\theta}$ together with $\Omega_k$, $\Omega_q$ and $\Omega_v$ are updated by maximizing (resp. minimizing) the agreement of the aligned local representations of each pair of examples from the same class (resp. different classes).
    }
    \label{pip}
\end{figure*}

\begin{figure}[t]
    \centering
    \includegraphics[width=0.47\textwidth]{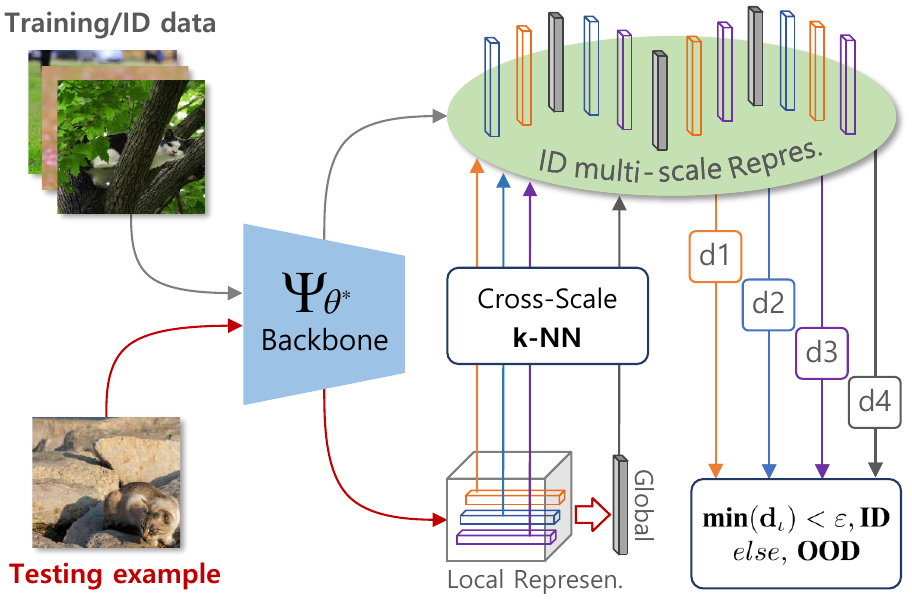}
    \caption{An overview of \textit{test-time} OOD detection with \textbf{Cross-scale Decision} ({$\mathtt{CSD}$}) where the most discriminative multi-scale (i.e., both global and local) representations are explored to distinguish ID/OOD examples more faithfully.
    }
    \label{test}
\end{figure}

In this work, we answer the above question by proposing \textbf{M}ulti-scale \textbf{O}OD \textbf{DE}tection (\textbf{MODE}), a first framework that leverages multi-scale (i.e., both global and local) representations for OOD detection.
Specifically, as illustrated in Fig. \ref{mtv}, we first find that existing models pretrained by off-the-shelf cross-entropy (CE) or contrastive learning (CL) losses are incompetent to capture valuable local representations for MODE, due to the \textit{scale-discrepancy} between the ID training and OOD detection procedures. 
To address this issue, we propose Attention-based Local PropAgation ($\mathtt{ALPA}$), a trainable
objective that encourages the mining of locally discriminative representations from images during ID training.
As shown in Fig. \ref{pip}, $\mathtt{ALPA}$ exploits contrastive representation learning to promote general-purpose visual information that captures richer and more flexible  representations for recognizing ID/OOD data. 
Yet, instead of directly using global representations to maximize/minimize the agreement of pairwise examples, $\mathtt{ALPA}$ adopts a cross-attention mechanism to align and highlight the local regions of the target objects for each pair of examples, making the extracted local representations more discriminative.
In test-time OOD detection, a Cross-Scale Decision (\textbf{$\mathtt{CSD}$}) function is further devised for MODE, where the most discriminative multi-scale representations are explored to distinguish ID/OOD examples more faithfully, as shown in Fig. \ref{test}.

\keypoint{Flexibility and Strong Performance.} 
The proposed MODE is orthogonal to the ID training procedure, as well as models pretrained with different fashions. 
More specifically, MODE can not only take $\mathtt{ALPA}$ as a plugin to regularize ID training losses, but also directly leverage it to finetune existing pre-trained models in an end-to-end manner. 
We demonstrate the effectiveness and flexibility of MODE on a broad spectrum of baseline methods applied to various network structures.   
Remarkably, our MODE establishes new state-of-the-art performance on several benchmarks, on average outperforming the previous best scheme KNN \cite{sun2022out} by up to \underline{\textbf{19.24\%}} in FPR, and  \underline{\textbf{2.77\%}} in AUROC (see Table \ref{sota1}).
What's more, when MODE performs test-time OOD detection based only on \underline{\textbf{5\%}} ID training data, it still exhibits superior performance than the strong competitor KNN (which relies on \underline{\textbf{100\%}} ID training examples), outperforming KNN by \underline{\textbf{6.08\%}} in FPR, \underline{\textbf{0.68\%}} in AUROC (see Table \ref{time}).

 \keypoint{Contributions.} To sum up, our contributions are fourfold.
 \begin{itemize}
  \item We propose MODE, a first framework that takes advantage of multi-scale (i.e., both global and local) representations for OOD detection.
   
  \item During ID training, we develop $\mathtt{ALPA}$, an end-to-end, plug-and-play, and cross-attention based learning objective tailored for encouraging locally discriminative representations for MODE.

   \item During test-time OOD detection, we devise \textbf{$\mathtt{CSD}$}, a simple, effective and multi-scale representations based ID-OOD decision function for MODE.

   \item Comprehensive experimental results on several benchmark datasets demonstrate the effectiveness and flexibility of MODE. Remarkably, our MODE achieves significantly better performance than state-of-the-art methods. 


\end{itemize}

\section{Related Work}
In  this section, we briefly review previous research closely related to our work, including out-of-distribution (OOD) detection, distance-based OOD detection, representation learning for OOD detection, and part-based visual correspondence.

\keypoint{Out-of-distribution Detection.}
Out-of-distribution (OOD) detection, a.k.a. \textit{outlier} detection \cite{lu2017unsupervised,zhao2017consensus}, 
\textit{anomaly} detection \cite{wu2021learning,sabokrou2017deep} or \textit{novelty} detection \cite{lo2022adversarially,hu2021global}, 
aims to recognize unknown inputs from the open world to prevent unpredictable risks.  
The vast majority of previous works are \textit{test-time} approaches that rely on the output softmax confidence score of a pretrained model to safeguard against OOD inputs. 
The insight beneath this line of works is that incoming examples with lower output softmax confidence scores are more likely from OOD \cite{liang2017enhancing,liu2020energy}.
Effective test-time scoring functions include OpenMax \cite{bendale2015towards}, MSP \cite{hendrycks2016baseline}, LogitNorm \cite{wei2022mitigating}, DICE \cite{sun2022dice}, Energy \cite{liu2020energy}, ODIN \cite{liang2017enhancing} and etc.
\textcolor{black}{In the recent work \cite{ahn2023line}, a simple yet effective test-time approach named LINE is proposed. By leveraging important neurons for post-hoc OOD detection, LINE yields remarkable test-time OOD detection performance.}
While the results are impressive, it has been demonstrated that well-performed models can produce arbitrarily high softmax confidence for inputs far away from the training data \cite{nguyen2015deep}.
Moreover, most of those test-time OOD methods often consider the development of effective OOD decision functions alone, our proposed MODE framework in this work considers training-time representation learning and test-time OOD detection, simultaneously.


\keypoint{Distance-based OOD Detection.}
The core concept of the distance-based OOD detection is to calculate a distance metric between the input examples and the training data.
Testing examples are recognised as OOD (resp. ID) data if they are relatively far away from (resp. close to) training examples of ID classes.
With the recent advances in representation learning, various kinds of distance-based OOD detection algorithms have been employed.
Among those methods, the Mahalanobis distance-based methods possess remarkable performance \cite{lee2018simple, ren2021simple}. 
However, the success of those methods are established on a strong distributional assumption of the underlying representation space, which may not always held in reality. 
To address this limitation, Sun et al. proposed KNN \cite{sun2022out}, a first study exploring the effectiveness of using a \textit{k}-nearest neighbor search over the penultimate layer representations for OOD detection. 
In contrast to the Mahalanobis distance-based methods, KNN \cite{sun2022out} does not impose any distributional assumptions on the underlying representation space, which is more simple, flexible and effective. 
\textcolor{black}{In \cite{ming2023exploit}, a novel representation learning framework coined CIDER is presented to exploit hyperspherical embeddings for distance-based OOD detection.
Recently, utilizing large vision-language pre-trained models like CLIP \cite{radford2021clip} for multi-modal downstream tasks has achieved remarkable success.
By matching visual features with textual class prototypes in the CLIP model, an effective test-time method coined MCM is proposed for distance-based OOD detection in \cite{ming2022delving}.}
Despite the encouraging advantages of distance-based OOD detection, we observe that the background clutter as well as the large intra-class variation may drive the image-level representations from the same ID class far apart in a given representation space. 
As a result, it becomes more difficult to correctly distinguish ID/OOD examples based only on the pairwise distances calculated from global image representations. 
Moreover, it has been widely demonstrated that a global average pooled image representation can destroy image structures and result in the compromise of a substantial amount of discriminative local representations of the target objects \cite{li2019revisiting,zhang2020deepemd}.
In this work, for the first time, we exploit both global visual information and local region details from images to calculate the distance between each pair of examples for maximally benefiting distance-based OOD detection.

\keypoint{Representation Learning for OOD Detection.}
A good deal of methods have attempted to improve the compactness of intra-class examples during the ID training stage, so as to achieve better test-time OOD detection performance \cite{tack2020csi,bai2021decaug,nandy2020towards}. 
Contrastive representation learning \cite{cole2022does,zhang2022progressive,zhang2022free,fu2022generalized,myj,zeng2022video} that targets learning a discriminative representation space where positive samples are aligned while negative ones are dispersed, has been shown to improve OOD detection \cite{sehwagssd,tack2020csi,zhou2022knn}. 
In particular, Tack et al. \cite{tack2020csi} proposed a scheme named Contrasting Shifted Instances (CSI) to learn a representation well-suited for novelty detection.
In \cite{sehwagssd}, authors present an effective outlier detector based on unlabeled ID data along with the self-supervised representation learning technique.
Recent studies \cite{vaze2021open,fort2021exploring} also revealed that improving the closed-set (i.e. ID) classification accuracy is the key to further boosting OOD detection performance.
Another promising line of work improves ID training by conducting training-time regularization \cite{liu2020energy,ming2022poem,du2022towards}. 
Most of those regularization approaches, however, require the availability of abundant simulating OOD data, which may not held in practice.
Surprisingly, the obtained quantitive and qualitative results reveal that relying only on the ID training data, our devised loss function $\mathtt{ALPA}$ can shape the distributions of different classes to be more compact for benefiting both OOD detection and ID classification tasks.

\keypoint{\textcolor{black}{Attention-based Local Feature Alignment.}}
\textcolor{black}{Local feature alignment \cite{halimi2019unsupervised,kim2017fcss,lifchitz2019dense} has emerged as a powerful paradigm enabling meaningful representations by matching local features of images (or image-text pairs), and has achieved great success in a wide spectrum of tasks, such as domain adaptation  \cite{xu2021cdtrans,zuo2021attention},
image-text matching  \cite{lee2018stacked,zhang2022negative},
few-shot learning \cite{li2019revisiting,ye2020feat,zhang2023deta}.
Among those methods, the idea of utilizing cross-attention to enhance feature alignments has been extensively studied. 
Particularly, CDTrans \cite{xu2021cdtrans} applies cross-attention and self-attention for source-target domain alignment to learn discriminative domain-invariant and domain-specific features simultaneously. 
SCAN \cite{lee2018stacked} highlights the alignment of image regions and words in a sentence in cross-attention modules to learn modality-invariant features.
FEAT \cite{ye2020feat} adapts the image features produced by deep convolution neural networks (CNNs) to the target few-shot task with a set-to-set function (i.e., Transformer \cite{dosovitskiy2020vit}), yielding discriminative and informative features.
Different from those works that leverage task-specific supervision to encourage the interaction between local features, the devised $\mathtt{ALPA}$ formulates the learning objective as a contrastive loss, where the cross-attention module takes the output dense features of CNNs as input to maximize (resp. minimize) the agreement of each pair of samples from the same ID class (resp. different ID classes). 
In addition, the goal of most of those works is to learn a shared feature space to align features from different domains \cite{xu2021cdtrans} (or modalities\cite{lee2018stacked}), while our ALPA aims to learn a discriminative feature space where a suitable threshold or compact decision boundary can be established to distinguish ID/OOD data accurately. 
To the best of our knowledge, this work is the first to use the idea of attention-based local feature alignment to promote locally discriminative representations in OOD detection.}

\keypoint{\textcolor{black}{Multi-scale Representation Learning.}}
\textcolor{black}{Multi-scale representations are of great importance to plenty of vision tasks such as classification \cite{chen2021crossvit,jiao2021multi}, retrieval \cite{gao2020fashion,bai2021multi} and detection \cite{gao2019res2net,gao2021hierarchical}, significantly boosting the performance achieved on single-scale (i.e., global) representations in those fields. 
Unlike most works in those fields that use multi-scale representations to recognize ID categories, in this work we for the first time leverage multi-scale representations to enable better ID-OOD separability in OOD detection, which is more challenging due to the following reasons.
On the one hand, relying only on the training data of ID categories, the learned multi-scale representations may not be generalizable enough to recognize parts, objects, and their surrounding context of OOD data. 
On the other hand, the sample space of potential OOD data can be prohibitively large, even severely overlapped with the sample space of ID categories \cite{ming2022poem,du2022vos}, making it difficult to establish a decision boundary on the extracted multi-scale representations of ID categories and OOD data at test time.
}


\section{Methodology}
In this section, we elaborate on our MODE framework. 
Before that, we introduce some important preliminaries. 

\subsection{Preliminaries}

When dealing with supervised multi-class classification, we typically denote $\mathcal{X}$, $\mathcal{Y}$ as the input, output space, respectively. 
Let $P$ be a distribution over $\mathcal{X}\times \mathcal{Y}$, and $f: \mathcal{X} \mapsto \mathbb{R}^{|\mathcal{Y}|}$ be a neural network that takes input the examples drawn from $P$ to output a logit vector, which is then used to predict the label of an input example. 
Denote $\mathbb{D}^{\mathbf{\textbf{in}}}={\{(\mathbf{\textbf{x}}_{i},{y}_{i})\}^{s}_{i=1}}$ as the marginal distribution of $P$ for $\mathcal{X}$, which represents the distribution of in-distribution (ID) data. 
During test-time OOD detection, the environment can present a distribution $\mathbb{D}^{\mathbf{\textbf{out}}}$ over $\mathcal{X}$ of OOD data, whose label space $\mathcal{Y}^{\mathbf{\textbf{out}}}$  s.t. ${\mathcal{Y}^{\mathbf{\textbf{in}}}}\bigcap{\mathcal{Y}^{\mathbf{\textbf{out}}}}=\phi$.  

\keypoint{Out-of-distribution Detection.} 
Essentially, OOD detection can be viewed as a binary classification task, where the goal is to reject the “unknown” inputs to prevent any potential risk. 
More specifically, to determine whether an example $\mathbf{\textbf{x}}\in\mathcal{X}$ belongs to $\mathbb{D}^{\mathbf{\textbf{in}}}$ or not (i.e. $\mathbb{D}^{\mathbf{\textbf{out}}}$), the decision function can be made via a level set estimation:
\begin{equation}
{{\Gamma}_{\varepsilon}}(\mathbf{\textbf{x}})=\left\{ \begin{matrix}
   \mathbf{{ID}}  & S(\mathbf{\textbf{x}}) \ge \varepsilon  \\
   \mathbf{{OOD}} & S(\mathbf{\textbf{x}}) < \varepsilon   \\
\end{matrix} \right. ,
\label{e1}
\end{equation}
where the input example $\mathbf{\textbf{x}}$ is classified as ID (resp. OOD) if its obtained score $S(\mathbf{\textbf{x}})$ is higher (resp. lower) than the threshold $\varepsilon$.
In practice, $\varepsilon$ is typically selected so that a high fraction of ID data (e.g. 95\%) is
correctly classified.

\keypoint{KNN-based OOD Detection.} 
Recent advances in representation learning give rise to distance-based OOD detection that represents image data in an appropriate representation space and leverages a distance function to decide whether testing examples are ID/OOD according to their relative distances to the seen examples of ID classes. In particular,
Sun et al. proposed KNN \cite{sun2022out} that established state-of-the-art performance using a $k$-nearest neighbor (coined $k$-NN in the following) search over global image representations for OOD detection.

Let $\Psi_{\theta}$ be a feature backbone (parameterized by $\theta$) mapping the input $\mathbf{\textbf{x}}$ to a global average pooled representation $\mathbf{\textbf{g}} \in \mathbb{R}^{E}$.
KNN-based OOD detection normalizes the global representation $\mathbf{\textbf{z}}=\mathbf{\textbf{g}}/|| \mathbf{\textbf{g}}||_2$ for distance calculation.
Before testing an example $\tilde{\mathbf{\textbf{z}}}$, we first obtain the representation collection of ID training data, denoted as $\mathbb{S}=(\mathbf{\textbf{z}}_1, ..., \mathbf{\textbf{z}}_s)$.
During test-time OOD detection, we calculate the Euclidean distances $||\mathbf{\textbf{z}}_i-\tilde{\mathbf{\textbf{z}}}||_2$ w.r.t. representations $\mathbf{\textbf{z}}_i \in \mathbb{S}$.
Denote the reordered ID data as $\mathbb{S}^{\prime}=(\mathbf{\textbf{z}}_{(1)}, ..., \mathbf{\textbf{z}}_{(s)})$, the decision function for KNN-based OOD detection takes the form of
\begin{equation}
{{\Gamma}_{\varepsilon}}(\tilde{\mathbf{\textbf{z}}}; k)=\left\{ \begin{matrix}
   \mathbf{{ID}}  &  r_k(\tilde{\mathbf{\textbf{z}}})  < \varepsilon  \\
   \mathbf{{OOD}} & r_k(\tilde{\mathbf{\textbf{z}}}) \ge \varepsilon   \\
\end{matrix} \right.,
\end{equation}
where $r_k(\tilde{\mathbf{\textbf{z}}})=||\mathbf{\textbf{z}}_{(k)}-\tilde{\mathbf{\textbf{z}}}||_2$ indicates the $k$-th nearest neighbor.
The threshold $\varepsilon$ does not depend on OOD data, and can be selected when a large proportion of of ID data (e.g. 95\%) is correctly classified in practice.


\keypoint{Contrastive Representation Learning.} 
\textcolor{black}{We take advantage of contrastive representation learning \cite{khosla2020supervised} to promote general-purpose visual information that captures richer and more flexible representations usable for recognizing ID/OOD data.}
Concretely, we first project the global representation of $\mathbf{\textbf{x}}$, $\mathbf{\textbf{g}}$, into a lower dimensional space with a projection head $h$, i.e., $h(\mathbf{\textbf{g}}) \in \mathbb{R}^{e}, e \ll E$. 
Let $\psi(h(\mathbf{\textbf{g}}_{i}),h(\mathbf{\textbf{g}}_{j}))$ be the cosine similarity of every pair of images in the projected space.
We sample a batch of $N$ pairs of images and labels from the training data of ID classes, and augment every image in the batch to obtain $2N$ labeled data points.
The loss function of supervised contrastive representation learning can therefore be expressed as
\begin{equation}
  {\mathcal{L}_{con}}=\sum_{i=1}^{2N}{\frac{1}{2{{N}_{{{y}_{i}}}}-1}\sum_{j=1}^{2N}{{{\mathbbm{1}}_{i\ne j}}\cdot {{\mathbbm{1}}_{{{y}_{i}}={{y}_{j}}}}\cdot {{\ell}_{ij}}}},
  \label{con}
\end{equation}
and we have
\begin{equation}
  {{\ell}_{ij}}=-\log \frac{\exp (\psi(h(\mathbf{\textbf{g}}_{i}), h(\mathbf{\textbf{g}}_{j})/\tau)}{\sum_{t=1}^{2N}{{\mathbbm{1}_{i\ne t}}}\cdot \exp(\psi(h(\mathbf{\textbf{g}}_{i}),h(\mathbf{\textbf{g}}_{t}))/\tau)},
\end{equation}
where $\mathbbm{1}$ is the indicator function, and $N_{y_{j}}$ is the number of the samples with the same label $y_{j}$, $\tau$ is a scalar temperature parameter. 
The above learning objective ${\mathcal{L}_{con}}$ introduces the label information to avoid pulling augmented views from the same class apart, enabling the mining of more discriminative and robust representations.


\subsection{Multi-scale OOD Detection (MODE)}
\label{smode}
Our goal in this work is to take advantage of multi-scale (i.e., both global and local) representations from images to distinguish ID/OOD examples more effectively.
Particularly, \textit{local representations} are the output feature maps before the final global average pooling layer of convolutional neural networks (CNNs). 
For an input image $\mathbf{\textbf{x}}$, we denote the obtained $HW$ $E$-dimensional local representations as $\mathbf{\textbf{L}} = {\Psi_{\theta}}(\mathbf{\textbf{x}}) \in \mathbb{R}^{HW\times E}$, and the global representation as $\mathbf{\textbf{g}} = \nu(\textbf{L}) \in \mathbb{R}^{E}$, where $\Psi_{\theta}$ denotes a feature backbone, and $\nu : \mathbb{R}^{HW\times E} \mapsto \mathbb{R}^{E}$ is an additional average pooling layer. 
The multi-scale representations for $\mathbf{\textbf{x}}$ thus can be expressed as $\mathbb{M}=\{\mathbf{\textbf{g}}, \mathbf{\textbf{L}}\}$.

Intuitively, we can directly borrow existing pretrained CNNs to generate multi-scale representations for MODE.
Unfortunately, due to the \textit{scale-discrepancy} between the ID training and OOD detection processes, models learned by off-the-shelf Cross-Entropy (CE) or Contrastive Learning (CL) losses are incompetent to capture discriminative local representations for recognizing OOD data, as demonstrated in Fig. \ref{mtv}.
This observation is also consistent with abundant empirical  evidence that an average pooled image representation can destroy image structures and lose a substantial amount of discriminative local representations of the target objects during training \cite{boiman2008defense,li2019revisiting,wang2021dense}. 
And once the model has been learned, those lost valuable local representations become difficult to recover.
Hence, this challenge begs one important question:
\begin{framed}
Can we develop a model-agnostic approach to encourage locally discriminative representations in ID training, so as to overcome the scale-discrepancy issue and benefit MODE during testing?
\end{framed}

{\keypoint{Attention-based Local Propagation ($\mathtt{ALPA}$).}}
Our solution to the above question is $\mathtt{ALPA}$, a trainable loss function tailored for mining discriminative local representations during ID training.
$\mathtt{ALPA}$ exploits contrastive representation learning to promote general-purpose visual information that captures richer and more flexible representations for recognizing ID/OOD data.
However, instead of leveraging the global representations ($\mathbf{\textbf{g}}_i$, $\mathbf{\textbf{g}}_j$) as in Eq. \ref{con}, we use the local representations ($\mathbf{\textbf{L}}_i$, $\mathbf{\textbf{L}}_j$) to compute the similarity/dissimilarity of each pair of inputs ($\mathbf{\textbf{x}}_i$, $\mathbf{\textbf{x}}_j$), as can be observed in Fig. \ref{pip}.

Concretely, our $\mathtt{ALPA}$ adopts a cross-attention mechanism to align and highlight the local regions of the target objects for each pair of examples, so as to extract more discriminative local representations. 
Following the design of Transformers \cite{vaswani2017attention}, the key $(\mathbf{\textbf{K}})$, value $(\mathbf{\textbf{V}})$ and query $(\mathbf{\textbf{Q}})$ are first generated for $\mathbf{\textbf{L}}$ using three independent linear maps: the key-head $\Omega_k: \mathbf{\textbf{L}} \mapsto \mathbb{R}^{HW\times e_k}$, the value-head $\Omega_v: \mathbf{\textbf{L}} \mapsto \mathbb{R}^{HW\times e_v}$, the query-head $\Omega_q: \mathbf{\textbf{L}} \mapsto \mathbb{R}^{HW\times e_q}$, respectively.
Note that we set ${e} = e_k = e_v = e_q$ for simpler illustration.
Let $\mathbf{\textbf{L}}_i$ and $\mathbf{\textbf{L}}_j$ be the local representations of two pairwise examples $i$ and $j$ respectively. 
Our goal is to calculate the aligned values of $i$ w.r.t. $j$, denoted as $\mathbf{\textbf{V}}_{i|j}$.
To this end, we first use the key $\mathbf{\textbf{K}}_i$ and query $\mathbf{\textbf{Q}}_j$ to determine the attention weights $\mathbf{\textbf{a}}_{ij} \in \mathbb{R}^{HW\times HW}$, by which we can obtain $\mathbf{\textbf{V}}_{i|j}$:
\begin{equation}
\mathbf{\textbf{V}}_{i|j}=\mathbf{\textbf{a}}_{ij} \mathbf{\textbf{V}}_{i}, \, \, \, \, \mathbf{\textbf{a}}_{ij}=\mathbf{{softmax}} \left( \frac{\mathbf{\textbf{Q}}_{j}\mathbf{\textbf{K}}_{i}^{\top}}{\sqrt{e}} \right).
\end{equation}
In the same way, we can calculate $\mathbf{\textbf{V}}_{j|i}$ by aligning the value of $j$ w.r.t. $i$ using the key $\mathbf{\textbf{K}}_j$ and the query $\mathbf{\textbf{Q}}_i$.



Therefore, for each pair values $\mathbf{\textbf{V}}_i$ and $\mathbf{\textbf{V}}_j$, we can determine their two aligned formulations, i.e., $\mathbf{\textbf{V}}_{i|j}$ and $\mathbf{\textbf{V}}_{j|i}$, by using the aforementioned cross-attention mechanism.
We then conduct an $l_2$ normalization on each representation vector $\mathbf{\textbf{V}}_{*}^{l}$ of all these values, $l=1,2,...,HW$.
The similarity between $\mathbf{\textbf{L}}_i$ and $\mathbf{\textbf{L}}_j$ can thus be expressed as
\begin{equation}
\mathbf{sim}(\mathbf{\textbf{L}}_i,\mathbf{\textbf{L}}_j)=\frac{1}{HW}\sum\limits_{l=1}^{HW}{\left[ {{\left(\mathbf{\textbf{V}}_{i}^{l} \right)}^{\top}}\mathbf{\textbf{V}}_{j|i}^{l}+{{\left(\mathbf{\textbf{V}}_{j}^{l} \right)}^{\top}}\mathbf{\textbf{V}}_{i|j}^{l} \right]}
\end{equation}
Finally, similar to Eq. \ref{con}, we can define the loss function of our $\mathtt{ALPA}$  as follows:
\begin{equation}
    \mathcal{L}_{alpa}=\sum_{i=1}^{2N}{\frac{1}{2{{N}_{{{y}_{i}}}}-1}\sum_{j=1}^{2N}{{{\mathbbm{1}}_{i\ne j}}\cdot {{\mathbbm{1}}_{{{y}_{i}}={{y}_{j}}}}\cdot {{\ell}_{ij}}}},
  \label{lalpa}
\end{equation}
where
\begin{equation}
  {{\ell}_{ij}}=-\log \frac{\exp (\mathbf{sim}(\mathbf{\textbf{L}}_i,\mathbf{\textbf{L}}_j)/\tau)}{\sum_{t=1}^{2N}{{\mathbbm{1}_{i\ne t}}}\cdot \exp(\mathbf{sim}(\mathbf{\textbf{L}}_i,\mathbf{\textbf{L}}_t)/\tau)}.
\end{equation}


Essentially, the objective $\mathcal{L}_{alpa}$ in Eq. \ref{lalpa} breaks the boundaries of training examples from the same class, and makes use of their local representations collectively to provide richer and more flexible representations for every class. 
In addition, by leveraging a cross-attention mechanism to align and highlight the local representations of the target objects for each pair of examples, the adverse effects of background clutter and intra-class variation can be significantly alleviated.
In this vein, the learned local representations are more transferable and capture more critical patterns outside the ID classes.

\keypoint{Remark \textbf{\textcircled{1}}: $\mathtt{ALPA}$ As A Plugin.} 
In practice, the developed $\mathtt{ALPA}$ can be utilized as a plugin to encourage locally discriminative representations by \textbf{i}) regularizing the ID training procedure, coined $\mathtt{ALPA}$-$\mathtt{train}$, or \textbf{ii}) directly finetuning models pretrained with different fashions, in an end-to-end manner, dubbed $\mathtt{ALPA}$-$\mathtt{finetune}$.
To sum up, the $\mathtt{ALPA}$-enhanced, ID training objective can take the form of
\begin{equation}
\mathcal{L}=\left\{ \begin{matrix}
   \mathcal{L}_{alpa}  &  \mathtt{ALPA}$-$\mathtt{finetune} \\
   \mathcal{L}_{base}+ \lambda\mathcal{L}_{alpa} & \mathtt{ALPA}$-$\mathtt{train}  \\
\end{matrix} \right.,
\label{lll}
\end{equation}
where $\mathcal{L}_{base}$ indicates the learning objective of the existing ID training (or representation learning) procedure, $\lambda$ is a balance weight controlling the contribution of $\mathtt{ALPA}$ for $\mathtt{ALPA}$-$\mathtt{train}$.

\keypoint{{Remark \textbf{\textcircled{2}}: Key Differences Between ALPA and DenseCL.}} 
The work most closely related to ALPA is DenseCL \cite{wang2021dense}, which implements contrastive representation learning by formulating a dense-feature-level contrastive loss based on different views of images.
We highlight the key differences between $\mathtt{ALPA}$ and DenseCL as follows.
On the one hand, DenseCL designs the loss function in an unsupervised learning setting, while our $\mathtt{ALPA}$ leverages label information to avoid pulling augmented views from the same class apart.
On the other hand, DenseCL uses an identical $1\times 1$ convolution layer as a projection head to generate lower-dimensional dense feature vectors for individual examples, while our $\mathtt{ALPA}$ exploits a cross-attention mechanism to highlight the object regions of pairwise examples, making the learned representations more discriminative and robust.
Nevertheless, DenseCL \cite{wang2021dense} does bring some inspiration to our method.

\keypoint{Remark \textbf{\textcircled{3}}: Time Complexity of $\mathtt{ALPA}$.} 
The time complexity of our $\mathtt{ALPA}$ is $\mathcal{O}(\Pi^2)$, where $\Pi=N\times HW \times e$.
Thus, we can \textbf{i)} reduce the batch size $N$, \textbf{ii)} apply an extra average pooling step on $\mathbf{\textbf{L}} \in \mathbb{R}^{HW\times E}$ (to reduce $HW$), and \textbf{iii)} set a smaller dimensionality $e$ in attention heads to avoid excessive computational cost in practice.

\keypoint{OOD Detection with Cross-scale Decision ({$\mathtt{CSD}$}).}
In the ID training procedure, we propose $\mathtt{ALPA}$, an end-to-end, plug-and-play, and cross-attention based loss function for encouraging locally discriminative representations for MODE.
To maximally benefit test-time OOD detection, we develop a Cross-Scale Decision ({$\mathtt{CSD}$}) function to distinguish ID/OOD examples more faithfully, relying on the relative distances of the most discriminative multi-scale representations.

Mathematically, let $\Psi_{\theta^*}$ be the $\mathtt{ALPA}$-enhanced feature backbone. 
We first apply $\Psi_{\theta^*}$ and $\nu()$ to produce the multi-scale representations for the training data $\mathbb{D}^{\mathbf{\textbf{tr}}}=\{\textbf{x}_{1}, ..., \textbf{x}_{s}\}$, denoted as 
$\mathbb{M}^{\mathbf{\textbf{tr}}}=\{\mathbf{\textbf{g}}_1, ..., \mathbf{\textbf{g}}_s, \textbf{L}_1^{1}, ..., \textbf{L}_1^{HW}, ..., \textbf{L}_s^{1}, ..., \textbf{L}_s^{HW}\}=\{\mathbf{\textbf{m}}_1^{\mathbf{\textbf{tr}}}, ..., \mathbf{\textbf{m}}^{\mathbf{\textbf{tr}}}_{(HW+1)\times s}\}$, where $\mathbf{\textbf{m}}^{\mathbf{\textbf{tr}}}_* \in \mathbb{R}^{E}$.
Similarly, denote the multi-scale representations of $i$-th testing example as $\mathbb{M}_i^{\mathbf{\textbf{test}}}=\{\textbf{m}_{1}^{\mathbf{\textbf{test}}}, ..., \textbf{m}_{HW+1}^{\mathbf{\textbf{test}}}\}$. 
For each $\textbf{m}_{i}^{\mathbf{\textbf{test}}}$, we search over $\mathbb{M}^{\mathbf{\textbf{tr}}}$ to determine its distance to the \textit{k}-th nearest neighbor in a normalized representation space (as in KNN \cite{sun2022out}), denoted as $r_k(\textbf{m}_{i}^{\mathbf{\textbf{test}}})$. 
The decision function of \textbf{$\mathtt{CSD}$} for distinguish ID/OOD data takes the form of
\begin{equation}
{{\Gamma}_{\varepsilon}}(\mathbb{K}_i^{(k)})=\left\{ \begin{matrix}
   \mathbf{{ID}}  &  \mathbf{min}(\mathbb{K}_i^{(k)})   < \varepsilon  \\
   \mathbf{{OOD}} & \mathbf{min}(\mathbb{K}_i^{(k)}) \ge \varepsilon   \\
\end{matrix} \right. ,
\label{csd}
\end{equation}
where $\mathbb{K}_i^{(k)}=\{r_k(\textbf{m}_{1}^{\mathbf{\textbf{test}}}), ..., r_k(\textbf{m}_{HW+1}^{\mathbf{\textbf{test}}})\}$, and $r_k(\textbf{m}_{i}^{\mathbf{\textbf{test}}})$ be the calculate the Euclidean distance between each representation in $\mathbb{M}_i^{\mathbf{\textbf{test}}}$ and its searched \textit{k}-th nearest neighbor in $\mathbb{M}^{\mathbf{\textbf{tr}}}$.
In like manner, the threshold $\varepsilon$ can be selected when a large proportion of of ID data (e.g. 95\%) is correctly classified.

\begin{table*}[]
\caption{Comparison with state-of-the-arts on common benchmarks.  
$^{*}$ and $^{\dagger}$ indicate the baseline model is pretrained with the vanilla
cross-entropy (\textbf{CE}) loss and contrastive learning (\textbf{CL}) loss, respectively. 
$ ^{\uparrow} $ (resp. $ ^{\downarrow} $) indicates larger (resp. smaller) is better. 
\textbf{MODE-T} = $\mathtt{ALPA}$-$\mathtt{train}$ + $\mathtt{CSD}$, 
\textbf{MODE-F} = $\mathtt{ALPA}$-$\mathtt{finetune}$ + $\mathtt{CSD}$.
The best (resp. second best) avg results are \textbf{bold} (resp. \underline{underlined}).
}
\tabcolsep 0.073in
\begin{tabular}{cccccccccccccc}
\toprule
\multicolumn{2}{c}{\multirow{2}{*}{\textbf{Method}}}  & \multicolumn{2}{c}{SVHN}      & \multicolumn{2}{c}{Places365} & \multicolumn{2}{c}{LSUN}      & \multicolumn{2}{c}{iSUN}      & \multicolumn{2}{c}{Texture}   & \multicolumn{2}{c}{\textbf{Average}}   \\
                 &                  & \textbf{FPR$\downarrow$} & \textbf{AUROC$\uparrow$} & \textbf{FPR$\downarrow$} & \textbf{AUROC$\uparrow$} & \textbf{FPR$\downarrow$} & \textbf{AUROC$\uparrow$} & \textbf{FPR$\downarrow$} & \textbf{AUROC$\uparrow$} & \textbf{FPR$\downarrow$} & \textbf{AUROC$\uparrow$} & \textbf{FPR$\downarrow$} & \textbf{AUROC$\uparrow$} \\
\midrule
\multicolumn{2}{c}{MSP\cite{hendrycks2016baseline}$^*$}      & 59.66        & 91.25          & 62.46        & 88.64          & 51.93        & 92.73          & 54.57        & 92.12          & 66.45        & 88.50          & 59.01        & 90.65          \\
\multicolumn{2}{c}{Energy\cite{liu2020energy}$^*$}          & 54.41        & 91.22          & 42.77        & 91.02          & 23.45        & 96.14          & 27.52        & 95.59          & 55.23        & 89.37          & 40.68        & 92.67          \\
\multicolumn{2}{c}{ODIN\cite{liang2017enhancing}$^*$}            & 20.93        & 95.55          & 63.04        & 86.57          & 31.92        & 94.82          & 33.17        & 94.65          & 56.40        & 86.21          & 41.09        & 91.56          \\
\multicolumn{2}{c}{GODIN\cite{hsu2020generalized}$^*$}           & 15.51        & 96.60          & 62.63        & 87.31          & 32.43        & 95.08          & 34.03        & 94.94          & 46.91        & 89.69          & 38.30        & 92.72          \\
\multicolumn{2}{c}{Mahalanobis\cite{lee2018simple}$^*$}     & 9.24         & 97.80          & 83.50        & 69.56          & 4.76         & 98.85          & 6.02         & 98.63          & 23.21        & 92.91          & 25.35  
& 91.55          \\     
\multicolumn{2}{c}{{{KNN}}\cite{sun2022out}$^*$}     & 24.53     &95.96     & 25.29   &  95.69   &25.55       & 95.26       &  27.57     &94.71        & 50.90       &  89.14    &  30.77     &   94.15     \\
\rowcolor{gray!20} \multicolumn{2}{c}{\textbf{MODE-T} (\textit{ours})$^*$}   &4.33 &97.67  &25.01 &95.45  &7.79  &98.46  &34.08  & 94.87 & 24.14 & 94.74  &\underline{19.07}  &\underline{96.24}  \\
\rowcolor{gray!20} \multicolumn{2}{c}{\textbf{MODE-F} (\textit{ours})$^*$}     & 1.00 & 99.79 & 29.58 & 94.39 &5.06  & 99.12 &32.55  &95.03  &19.56 &97.11   &\textbf{17.55}  &\textbf{97.09}  \\  
\midrule
\multicolumn{2}{c}{CE+SimCLR\cite{winkens2020contrastive}$^{\dagger}$}     & 6.98         & 99.22          & 54.39        & 86.70          & 64.53        & 85.60          & 59.62        & 86.78          & 16.77        & 96.56          & 40.46        & 90.97          \\
\multicolumn{2}{c}{CSI\cite{tack2020csi}$^{\dagger}$}         & 37.38        & 94.69          & 38.31        & 93.04          & 10.63        & 97.93          & 10.36        & 98.01          & 28.85        & 94.87          & 25.11        & 95.71          \\
\multicolumn{2}{c}{SSD\cite{sehwagssd}$^{\dagger}$}            & 2.47         & 99.51          & 22.05        & 95.57          & 10.56        & 97.83          & 28.44        & 95.67          & 9.27         & 98.35          & 14.56        & 97.38          \\
\multicolumn{2}{c}{ProxyAnchor\cite{kim2020proxy}$^{\dagger}$}     & 39.27        & 94.55          & 43.46        & 92.06          & 21.04        & 97.02          & 23.53        & 96.56          & 42.70        & 93.16          & 34.00        & 94.46          \\
\multicolumn{2}{c}{\textcolor{black}{LINE}\cite{ahn2023line}$^{\dagger}$}    &\textcolor{black}{5.46}&\textcolor{black}{98.37} &\textcolor{black}{23.98} &\textcolor{black}{94.71} &\textcolor{black}{4.49} &\textcolor{black}{98.43} &\textcolor{black}{22.11} &\textcolor{black}{96.72} &\textcolor{black}{9.12} &\textcolor{black}{97.86} &\textcolor{black}{{13.03}}  &\textcolor{black}{97.21}  \\
\multicolumn{2}{c}{\textcolor{black}{CIDER}\cite{ming2023exploit}$^{\dagger}$}    &\textcolor{black}{2.89}&\textcolor{black}{99.72} &\textcolor{black}{23.88} &\textcolor{black}{94.09} &\textcolor{black}{5.45} &\textcolor{black}{99.01} &\textcolor{black}{20.61} &\textcolor{black}{96.64} &\textcolor{black}{12.33} &\textcolor{black}{96.85} &\textcolor{black}{12.95} & \textcolor{black}{97.26} \\
\multicolumn{2}{c}{KNN\cite{sun2022out}$^{\dagger}$}    & 2.42       & 99.52    & 23.02    &95.36      & 1.78        & 99.48         &20.06         & 96.74         & 8.09        & 98.56       & \underline{11.07}          & \underline{97.93}          \\
\rowcolor{gray!20} \multicolumn{2}{c}{\textbf{MODE-T} (\textit{ours})$^{\dagger}$}   &1.24 &99.76  &24.51 &95.20  &3.83  &99.22  &24.41  & 96.22 &9.93 &98.31   &{12.78}  &{97.74}  \\
\rowcolor{gray!20} \multicolumn{2}{c}{\textbf{MODE-F} (\textit{ours})$^{\dagger}$} & {0.65}  &  {99.86}   &  {20.13}           &  {96.44}     & {2.15}   &    {99.31}  &   19.87   &  {97.59}   &     {8.46}   &  {98.22}           &  \textbf{10.05} &  {\textbf{98.42}}              \\ 
\bottomrule
\multicolumn{14}{c}{\multirow{2}{*}{\textbf{(a) CIFAR-10 (ID) with ResNet-18}}}    \\
\multicolumn{14}{c}{\multirow{2}{*}{}}   \\
\end{tabular}
\begin{tabular}{cccccccccccccc}
\toprule
\multicolumn{2}{c}{\multirow{2}{*}{\textbf{Method}}} & \multicolumn{2}{c}{SVHN}      & \multicolumn{2}{c}{Places365} & \multicolumn{2}{c}{LSUN}      & \multicolumn{2}{c}{iSUN}      & \multicolumn{2}{c}{Texture}   & \multicolumn{2}{c}{\textbf{Average}}   \\
                 &                  & \textbf{FPR$\downarrow$} & \textbf{AUROC$\uparrow$} & \textbf{FPR$\downarrow$} & \textbf{AUROC$\uparrow$} & \textbf{FPR$\downarrow$} & \textbf{AUROC$\uparrow$} & \textbf{FPR$\downarrow$} & \textbf{AUROC$\uparrow$} & \textbf{FPR$\downarrow$} & \textbf{AUROC$\uparrow$} & \textbf{FPR$\downarrow$} & \textbf{AUROC$\uparrow$} \\
\midrule
\multicolumn{2}{c}{MSP\cite{hendrycks2016baseline}$^*$}            & 78.89        & 79.80          & 84.38        & 74.21          & 83.47        & 75.28          & 84.61        & 74.51          & 86.51        & 72.53          & 83.12        & 75.27          \\
\multicolumn{2}{c}{ODIN\cite{liang2017enhancing}$^*$}            & 70.16        & 84.88          & 82.16        & 75.19          & 76.36        & 80.10          & 79.54        & 79.16          & 85.28        & 75.23          & 78.70        & 79.11          \\
\multicolumn{2}{c}{Mahalanobis\cite{lee2018simple}$^*$}     & 87.09        & 80.62          & 84.63        & 73.89          & 84.15        & 79.43          & 83.18        & 78.83          & 61.72        & 84.87          & 80.15        & 79.53          \\
\multicolumn{2}{c}{Energy\cite{liu2020energy}$^*$}          & 66.91        & 85.25          & 81.41        & 76.37          & 59.77        & 86.69          & 66.52        & 84.49          & 79.01        & 79.96          & 70.72        & 82.55          \\
\multicolumn{2}{c}{GODIN\cite{hsu2020generalized}$^*$}           & 74.64        & 84.03          & 89.13        & 68.96          & 93.33        & 67.22          & 94.25        & 65.26          & 86.52        & 69.39          & 87.57        & 70.97          \\
\multicolumn{2}{c}{LogitNorm\cite{wei2022mitigating}$^*$}       & 59.60        & 80.74          & 80.25        & 78.58          & 91.07        & 82.99          & 84.19        & 80.77          & 86.64        & 75.60          & 78.35        & 81.74          \\
\multicolumn{2}{c}{{{KNN}}\cite{sun2022out}$^*$}     & 29.08     & 93.90   & 87.50   & 72.35    &  87.97    & 74.11      &  91.62    &   80.55    &  47.66     &  87.44   &   66.77    &  81.67      \\
\rowcolor{gray!20} \multicolumn{2}{c}{\textbf{MODE-T} (\textit{ours})$^*$}        &27.50  &93.94  &59.63 &83.14  & 40.67 &75.68  &64.35  & 86.20 &52.71 & 87.59  & \underline{48.97} &\textbf{85.31}  \\  
\rowcolor{gray!20} \multicolumn{2}{c}{\textbf{MODE-F} (\textit{ours})$^*$}      & 24.13 &94.11  &61.95 &82.22  &34.76  &72.07  & 65.52 &86.48  &51.28 &87.30   &\textbf{47.53}  &\underline{84.44}  \\   
\midrule
\multicolumn{2}{c}{ProxyAnchor\cite{kim2020proxy}$^{\dagger}$}     & 87.21        & 82.43          & 70.10        & 79.84          & 37.19        & 91.68          & 70.01        & 84.96          & 65.64        & 84.99          & 66.03        & 84.78          \\
\multicolumn{2}{c}{CE+SimCLR\cite{winkens2020contrastive}$^{\dagger}$}     & 24.82        & 94.45          & 86.63        & 71.48          & 56.40        & 89.00          & 66.52        & 83.82          & 63.74        & 82.01          & 59.62        & 84.15          \\
\multicolumn{2}{c}{CSI\cite{tack2020csi}$^{\dagger}$}             & 44.53        & 92.65          & 79.08        & 76.27          & 75.58        & 83.78          & 76.62        & 84.98          & 61.61        & 86.47          & 67.48        & 84.83          \\
\multicolumn{2}{c}{SSD\cite{sehwagssd}$^{\dagger}$}            & 31.19        & 94.19          & 77.74        & 79.90          & 79.39        & 85.18          & 80.85        & 84.08          & 66.63        & 86.18          & 67.16        & 85.90          \\
\multicolumn{2}{c}{\textcolor{black}{LINE}\cite{ahn2023line}$^{\dagger}$}    &\textcolor{black}{30.45}&\textcolor{black}{92.21} &\textcolor{black}{79.54} &\textcolor{black}{77.67} &\textcolor{black}{50.70} &\textcolor{black}{89.01} &\textcolor{black}{67.84} &\textcolor{black}{83.64} &\textcolor{black}{54.60} &\textcolor{black}{89.37} &\textcolor{black}{56.62}  & \textcolor{black}{86.38} \\
\multicolumn{2}{c}{\textcolor{black}{CIDER}\cite{ming2023exploit}$^{\dagger}$}   &\textcolor{black}{23.09} &\textcolor{black}{95.16} &\textcolor{black}{79.63} &\textcolor{black}{73.43} &\textcolor{black}{16.16} &\textcolor{black}{96.33}&\textcolor{black}{71.68} &\textcolor{black}{82.98} &\textcolor{black}{43.87} &\textcolor{black}{90.42} &\textcolor{black}{\textbf{46.89}} & \textcolor{black}{\underline{87.67}} \\
\multicolumn{2}{c}{KNN\cite{sun2022out}$^{\dagger}$}            & 39.23        & 92.78          & 80.74        & 77.58          & 48.99        & 89.30     & 74.99        & 82.69          & 57.15        & 88.35          & 60.22        & 86.14    \\
\rowcolor{gray!20} \multicolumn{2}{c}{\textbf{MODE-T} (\textit{ours})$^{\dagger}$}      &29.37 &92.29  &73.91 &78.93  &48.16  &89.38  &75.96  &83.39  &51.37 &87.87   &{55.75}  &{86.38}  \\  
\rowcolor{gray!20} \multicolumn{2}{c}{\textbf{MODE-F} (\textit{ours})$^{\dagger}$}    &  {21.18}     &  {95.19}        & {67.88}              &     {80.24}    & {51.67}             &    {90.22}               &        {59.61}        &   {86.92}          &      {52.32}         &     {86.94}         &   \underline{50.53}             &          \textbf{87.90}                  \\ 
\bottomrule
\multicolumn{14}{c}{\multirow{2}{*}{\textbf{(b) CIFAR-100 (ID) with ResNet-34}}}    \\
\end{tabular}
\label{sota1}
\end{table*}

\begin{figure}[t]
    \centering
    \includegraphics[width=0.45\textwidth]{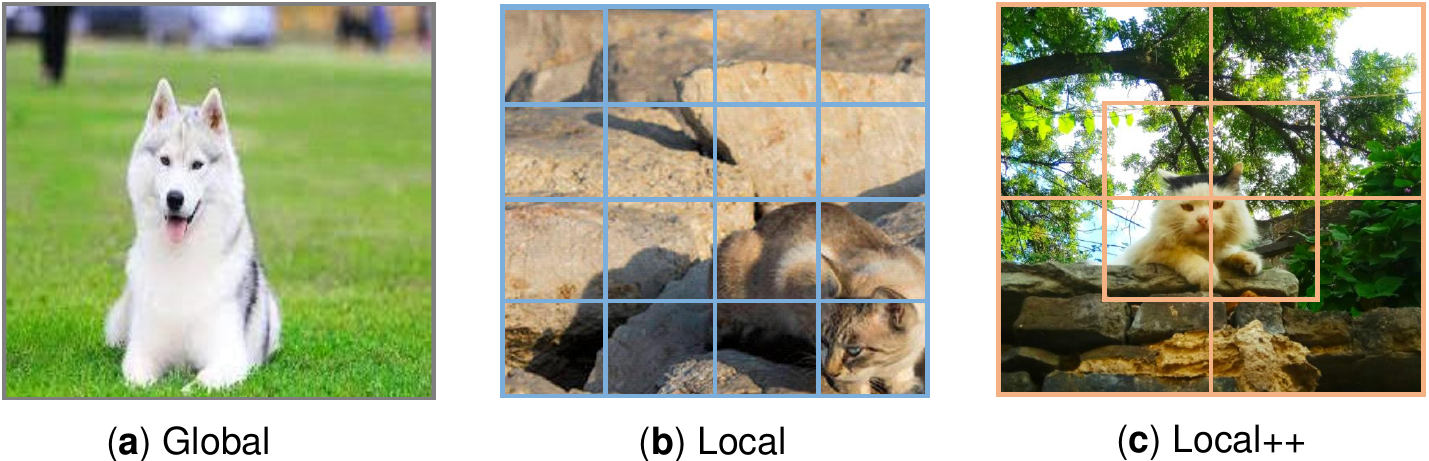}
    \caption{Illustrations of (\textbf{a}) {Global}, (\textbf{b}) {Local}, and (\textbf{c}) {Neighbor Aggregated} \textit{Local} (coined, {Local++}) representations. During test-time OOD detection, $\mathtt{CSD}$ leverages Global and Local++ representations for computational efficiency.}
    \label{nag}
\end{figure}

\keypoint{Remark \textbf{\textcircled{4}}: Strategies for Speeding Up $\mathtt{CSD}$.} 
In practice, to avoid excessive time cost, we follow KNN \cite{sun2022out} to \textbf{i}) store the multi-scale representations of all examples in a key-value map, and  
\textbf{ii}) use the library of $\mathbf{Faiss}$ \cite{johnson2019billion} for speeding up the $k$-NN search process. In concrete terms, we employ $\mathbf{faiss.IndexFlatL2}$ as the indexing scheme with Euclidean distance. 
Moreover, as illustrated in Fig. \ref{nag}, we further reduce the number of extracted local representations for every image from $HW$ to $HW/4+1$, by performing a \textit{neighbor aggregation} step (i.e., a $2\times 2$ average pooling step) on every four nearest local representations at different positions.
Quantitative analysis for the computational cost of our designed $\mathtt{CSD}$ at inference is presented in Section \ref{stime}.

\keypoint{Flexibility of our MODE Framework.}
Our proposed MODE framework is orthogonal to the ID training procedure, as well as models pretrained with different losses. 
In this work, we consider two versions of our MODE according to how $\mathtt{ALPA}$ promotes locally discriminative representations, i.e.,  
\textbf{MODE-T} = $\mathtt{ALPA}$-$\mathtt{train}$ + $\mathtt{CSD}$, 
\textbf{MODE-F} = $\mathtt{ALPA}$-$\mathtt{finetune}$ + $\mathtt{CSD}$.
In practice, we can flexibly decide whether to adopt MODE-T or MODE-F depending on the current training stage of the model, i.e., MODE-T if the model has not yet started training,  MODE-F if the model is already pretrained.

\section{Experiments.} 
In this section, we extensively test our proposed MODE on regularly used OOD benchmarks, feature backbones and evaluation metrics. In specific, we first scrutinize the effectiveness of our MODE on common benchmarks, then we move a step further to evaluate it on large-scale ImageNet benchmark. 
Ablation studies and visualization results are shown at the end.

\keypoint{Evaluation Metrics.}
We follow the widely-employed setup in the literature to use the following evaluation metrics:
{FPR} (a.k.a. FPR95) \cite{macedo2021entropic}: The false positive rate of OOD examples when the true positive rate of ID examples reaches 95\%.
{AUROC} \cite{hendrycks2016baseline}: The area under the curve of the receiver operating characteristic.
Note that both FPR and AUROC are utilized for testing the OOD detection performance, besides we don't need to manually tune the threshold $\varepsilon$ for FPR and AUROC at inference, as the two metrics can determine $\varepsilon$ according to the classification results of testing ID samples.
To investigate the ID training (or representation learning) performance we also introduce {ID ACC}: the classification accuracy of ID examples.

\subsection{Evaluation on Common Benchmarks}
\label{exp1}
\keypoint{Datasets.} 
Following the common benchmarks in OOD detection, we adopt CIFAR-10 and CIFAR-100 as in-distribution (ID) datasets, and they are spilled normally for conducting ID training. 
In test-time OOD detection, Textures \cite{cimpoi2014describing}, SVHN \cite{netzer2011reading}, Places365 \cite{zhou2017places}, LSUN-C \cite{yu2015lsun} and iSUN \cite{xu2015turkergaze} are used as OOD datasets for performance evaluation.
Specifically, Places365 consists of 365 scene categories of images, SVHN and iSUN are datasets with colored street numbers and a large scale of natural scenes. Besides, Textures is made up of images in the wild from 47 terms, and LSUN contains millions of images from 10 scenes and 20 object categories. 

\begin{table*}[]\centering
\tabcolsep 0.09in
\caption{Comparison with state-of-the-arts on large-scale ImageNet benchmark -- \textbf{ImageNet-1k (ID) with ResNet-50}.
$^{*}$ and $^{\dagger}$ indicate the baseline model is pretrained by vanilla
cross-entropy (\textbf{CE}) loss and contrastive learning (\textbf{CL}) loss, respectively. 
$ ^{\uparrow} $ (resp. $ ^{\downarrow} $) means larger (resp. smaller) is better. 
Following KNN \cite{sun2022out}, we only sample a tiny ratio ($1\%$) of training data for nearest neighbor search. \textbf{MODE-F} = $\mathtt{ALPA}$-$\mathtt{finetune}$ + $\mathtt{CSD}.$
The best (resp. second best) average results are \textbf{bold} (resp. \underline{underlined}).
}
\begin{tabular}{ccccccccccccc}
\toprule
\multicolumn{2}{c}{\multirow{2}{*}{\textbf{Method}}} & \multicolumn{2}{c}{iNaturalist}      & \multicolumn{2}{c}{SUN} & \multicolumn{2}{c}{Places365}      & \multicolumn{2}{c}{Texture}      & \multicolumn{2}{c}{Average}\\
                 &                  & \textbf{FPR$\downarrow$} & \textbf{AUROC$\uparrow$} & \textbf{FPR$\downarrow$} & \textbf{AUROC$\uparrow$} & \textbf{FPR$\downarrow$} & \textbf{AUROC$\uparrow$} & \textbf{FPR$\downarrow$} & \textbf{AUROC$\uparrow$} & \textbf{FPR$\downarrow$} & \textbf{AUROC$\uparrow$}  \\
\midrule
\multicolumn{2}{c}{MSP\cite{hendrycks2016baseline}$^*$}            & 54.99        & 87.74         & 70.83        & 80.86          & 73.99        &79.76          & 68.00        & 79.61         & 66.95        & 81.99         \\
\multicolumn{2}{c}{ODIN\cite{liang2017enhancing}$^*$}            & 47.66        & 89.66         & 60.15        & 84.59          & 67.89        & 81.78         & 50.23        & 85.62        & 56.48        &85.41      \\
\multicolumn{2}{c}{Energy\cite{liu2020energy}$^*$}          & 55.72        & 89.95          &59.26        & 85.89          & 64.92        & 82.86          & 53.72        & 85.99         & 58.41        & 86.17       \\
\multicolumn{2}{c}{GODIN\cite{hsu2020generalized}$^*$}           & 61.91        & 85.40          & 60.83       & 85.60          & 63.70        & 83.81          &77.85        &73.27        & 66.07        & 82.02       \\
\multicolumn{2}{c}{Mahalanobis\cite{lee2018simple}$^*$}       & 97.00        & 52.65          & 98.50        & 42.41         & 98.40        & 41.79          & 55.80        & 85.01          & 87.43 & 55.47   \\
\multicolumn{2}{c}{SSD\cite{sehwagssd}$^{\dagger}$}     &57.16        & 87.77         & 78.23     & 73.10          & 81.19        & 70.97          &36.37        & 88.52          & 63.24       & 80.09      \\
\multicolumn{2}{c}{{\textcolor{black}{LINE}}\cite{ahn2023line}$^{\dagger}$}     & \textcolor{black}{32.31}       & \textcolor{black}{92.51}          & \textcolor{black}{43.37}        &\textcolor{black}{90.50}          & \textcolor{black}{60.73}        & \textcolor{black}{84.81}          & \textcolor{black}{30.24}        & \textcolor{black}{91.97}          & \textcolor{black}{41.66}        & \textcolor{black}{89.95} \\
\multicolumn{2}{c}{{{KNN}}\cite{sun2022out}$^{\dagger}$}     & 30.83        & 94.72          & 48.91        &88.40          & 60.02        & 84.62          & 16.97        & 94.45          & \underline{39.18}        & \underline{90.55} \\
\rowcolor{gray!20} \multicolumn{2}{c}{\textbf{MODE-F} (\textit{ours})$^{\dagger}$}     & 29.11     & 96.46     & 46.39       & 89.73   &     54.38         &   87.80        & 15.65   &   94.95  & \textbf{36.38}     &  \textbf{92.24} \\
\bottomrule
\end{tabular}
\label{sota2}
\end{table*}

\keypoint{Implementation Details.}
We follow the common practice to use ResNet-18 as the feature backbone for CIFAR-10, and ResNet-34 for CIFAR-100 in our experiments.
We obtain the CE-trained and CL-trained models based on the open-source implementations of SupCE and SupCon in \cite{khosla2020supervised}\footnote{https://github.com/HobbitLong/SupContrast}, respectively.
We update the networks using stochastic gradient descent with momentum 0.9, and the weight decay is set to 0.0001.
In particular, the balance weight $\lambda$ in Eq. \ref{lll} is set to 1.0 for MODE-T (or $\mathtt{ALPA}$-$\mathtt{train}$). 
The initial learning rate $\eta$ is set to 0.1 for MODE-F (or $\mathtt{ALPA}$-$\mathtt{finetune}$).
The dimensionality $e$ in attention heads takes the value of 80. 
The temperature $\tau$ in $\mathcal{L}_{alpa}$ is  0.1. 
The $k$-NN hyperparameter $k$ is 50.
We found that the batch size $N$ has negligible effect on performance within a certain range, we therefore set $N=128$ to avoid excessive computational cost. 
We carefully adjust the critical hyperparamethers $\lambda$, $\eta$, and $k$ in our ablation studies.

\begin{figure}[t]
    \centering
    \includegraphics[width=0.48\textwidth]{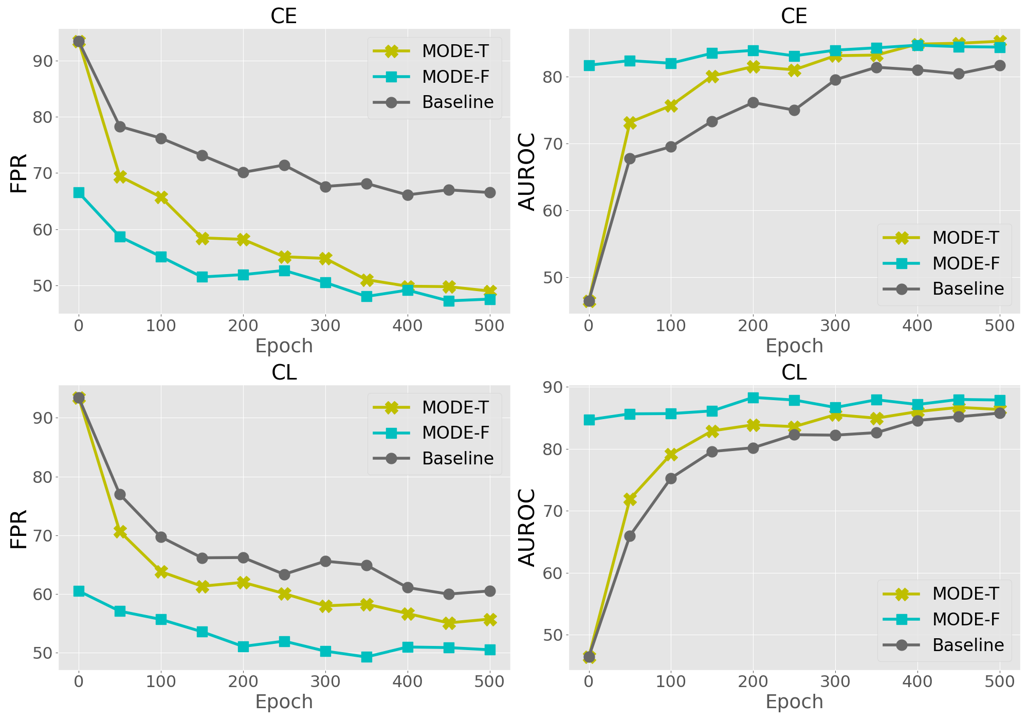}
    \caption{OOD detection performance curves of different approaches at various ID training epochs -- {CIFAR-100 (ID) with ResNet-34}. The reported results are the average across five common OOD datasets. For FPR (resp. AUROC), smaller (resp. higher) is better.}
    \label{curve}
\end{figure}

\keypoint{Experimental Results.}
The experimental results are reported in Table \ref{sota1}, where a broad spectrum of state-of-the-art OOD detection approaches are compared. 
Please visit \cite{sun2022out} for more details of those approaches.
In particular, we divide those approaches into two groups, depending on whether the pretrained model is learned by cross-entropy ({CE}) loss or contrastive learning ({CL}) loss.
From the reported results in the table, we highlight the following observations.
\underline{\textbf{First}}, OOD detection performance is substantially improved with our proposed MODE.
On average, with CE-trained (resp. CL-trained) models, our methods outperform the strong competitor KNN \cite{sun2022out} by a maximum of \textbf{19.24\%} (resp. \textbf{9.69\%}) in terms of FPR, and \textbf{2.77\%}  (resp. \textbf{1.76\%}) in terms of AUROC.
\underline{\textbf{Second}}, for the two versions of MODE, MODE-F (i.e. $\mathtt{ALPA}$-$\mathtt{finetune}$ + $\mathtt{CSD}$) outperforms MODE-T (i.e. $\mathtt{ALPA}$-$\mathtt{train}$ + $\mathtt{CSD}$) in the vast majority of cases, suggesting that our $\mathtt{ALPA}$-$\mathtt{finetune}$ does not raise the catastrophic forgetting problem -- overwriting the previously learned knowledge of pretrained models.
The OOD detection performance curves depicted in Fig. \ref{curve} further confirms this conclusion -- MODE-F continually improves the OOD detection performance of the pretrained baseline models.
\underline{\textbf{Third}}, the performance improvement of our MODE upon CE-trained methods is significantly better than that on CL-trained methods. 
One possible reason is that our devised $\mathtt{ALPA}$ as a variant of CL loss is able to complement vanilla CE loss to mine general-purpose visual information that captures richer
and more flexible representations for recognizing ID/OOD data. 
More qualitative results for this problem is systematically discussed in Section \ref{svis} and demonstrated in Fig. \ref{tsne}. 
In a nutshell, the achieved results in Table \ref{sota1} show that our MODE framework is agnostic to ID training losses, as well as models pretrained with different fashions. 

\subsection{Evaluation on Large-scale ImageNet Benchmark}
\label{DDDD}
\keypoint{Datasets.} 
We move a step further to demonstrate the effectiveness and flexibility of our method by evaluating it on a large-scale OOD detection task using ImageNet \cite{deng2009imagenet} as ID dataset. 
Following the common setup in ImageNet-based OOD detection \cite{huang2021mos,sun2022out}, we evaluate on four OOD datasets that are specifically the subsets of Textures \cite{cimpoi2014describing}, Places365 \cite{zhou2017places}, iNaturalist \cite{van2018inaturalist} and SUN \cite{xiao2010sun}, and without overlapping categories w.r.t. ImageNet.

\keypoint{Implementation Details.}
We use a ResNet-50 feature backbone for evaluation on the ID dataset ImageNet. 
Here, instead of meticulously training the backbone from scratch on ImageNet, we directly borrow the CL-trained ResNet-50 model from the public repository of KNN \cite{sun2022out}\footnote{https://github.com/deeplearning-wisc/knn-ood} for efficiency. 
Note that we only test our method on the CL-trained model, since the CE-trained model is not publicly available yet.
During ID training, we iteratively finetune the pretrained model using our $\mathtt{ALPA}$-$\mathtt{finetune}$ for 300 epochs, where the batch size $N=64$, and the initial learning rate $\eta=0.1$ with cosine scheduling.
Other hyperparameters are set the same as in Section \ref{exp1}.
In addition, following KNN \cite{sun2022out}, we sample a tiny ratio ($1\%$) of training data from ImageNet for nearest neighbor search during test-time OOD detection for our method and LINE \cite{ahn2023line}.

\keypoint{Experimental Results.}
The achieved OOD detection performance for different approaches over the four OOD datasets are reported in Table \ref{sota2}.
From the table, we have the following findings.
\underline{\textbf{First}}, our method (MODE-F) significantly outperforms those strong competitors across the four OOD datasets, and establishes new state-of-the-art results.
\underline{\textbf{Second}}, it is worth noting that SSD \cite{sehwagssd} obtains inferior performance to both KNN \cite{sun2022out} and our method.
This is probably due to that the increased data complexity of large-scale benchmarks makes the class-conditional Gaussian assumption less viable for effective OOD detection.
In contrast, KNN along with our method are distribution assumption-free therefore do not suffer from this limitation.
\underline{\textbf{Third}}, after finetuning the pretrained model (i.e. ResNet-50), our MODE-F (more concretely, the designed $\mathtt{CSD}$) randomly samples a tiny ratio ($1\%$) of training data for nearest neighbor search as in KNN. In this case, our method still consistently outperforms other competitors across all OOD datasets, revealing that the $\mathtt{ALPA}$-enhanced multi-scale representations are more informative and transferable.

\subsection{\textcolor{black}{Evaluation on Clean OOD Benchmarks}}
\textcolor{black}{As revealed in \cite{bitterwolf2023ninco}, most of widely-used OOD datasets are noisy: the test OOD data is mixed with a large proportion (by up to 50\% in some cases) of ID examples from ImageNet-1k.
To further show the effectiveness of our proposed method, we also compare our method with the strong baseline KNN \cite{sun2022out} on two \textit{clean} OOD datasets: OpenImage-O \cite{wang2022vim}, and NINCO \cite{bitterwolf2023ninco}. 
The obtained results are reported in Table \ref{clean}, where the experimental setup is the same as in Section \ref{DDDD}. From the results in the table, our method consistently outperforms the competitor KNN on the two  datasets.}

\begin{table}[]\centering
\tabcolsep 0.09in
\caption{{Evaluation on clean datasets (OpenImage-O, NINCO) -- \textbf{ImageNet-1k (ID) with ResNet-50}. 
$^{\dagger}$ the model is pretrained with contrastive loss. 
\textbf{MODE-F} = $\mathtt{ALPA}$-$\mathtt{finetune}$ + $\mathtt{CSD}$.}}
\begin{tabular}{cccccc}
\toprule
\multicolumn{2}{c}{\multirow{2}{*}{\textbf{\textcolor{black}{Method}}}} & \multicolumn{2}{c}{\textcolor{black}{OpenImage-O}} & \multicolumn{2}{c}{\textcolor{black}{NINCO}} \\
& & \textbf{\textcolor{black}{FPR$\downarrow$}} & \textbf{\textcolor{black}{AUROC$\uparrow$}} & \textbf{\textcolor{black}{FPR$\downarrow$}} & \textbf{\textcolor{black}{AUROC$\uparrow$}}  \\
\midrule
\multicolumn{2}{c}{\textcolor{black}{KNN}\cite{sun2022out}$^{\textcolor{black}{\dagger}}$}  &  \textcolor{black}{80.04}  & \textcolor{black}{69.88}  & \textcolor{black}{66.06}    & \textcolor{black}{84.11}\\
\rowcolor{gray!20} \multicolumn{2}{c}{\textcolor{black}{\textbf{MODE-F}} \textcolor{black}{(\textit{ours})}$^{\textcolor{black}{\dagger}}$} & \textbf{\textcolor{black}{75.73}}  & \textbf{\textcolor{black}{73.91}}  & \textbf{\textcolor{black}{64.42}} & \textbf{\textcolor{black}{85.07}} \\
\bottomrule
\end{tabular}
\label{clean}
\end{table}

\begin{table}[]
\tabcolsep 0.055in
\centering
\caption{Ablation study on the designed components of our MODE -- {CIFAR-100 (ID) with ResNet-34}. The results are the average across five common OOD datasets. $ ^{\uparrow} $ (resp. $ ^{\downarrow} $) indicates larger (resp. smaller) values are better.}
\begin{tabular}{ccccc}
\toprule
\textbf{ID Training}      & \textbf{ID ACC}$\uparrow$  & + \textbf{$\mathtt{CSD}$ (Testing)}      & \, \textbf{FPR}$\downarrow$ & \textbf{AUROC}$\uparrow$\\ 
\midrule
\multirow{2}{*}{\textbf{CE}}  & \multirow{2}{*}{73.23}   & \XSolidBrush   &66.77  &81.67\\
 &       & {\CheckmarkBold} &66.52   & 81.63    \\ 
\multirow{2}{*}{\textbf{+ $\mathtt{ALPA}$-$\mathtt{train}$}}   & \multirow{2}{*}{\textbf{75.52}}   & \XSolidBrush    & 52.31 & 83.94   \\  
 &       &\textcolor{red}{\CheckmarkBold} & \cellcolor{gray!20}48.97  &\cellcolor{gray!20}\textbf{85.31}   \\ 
\multirow{2}{*}{\textbf{+ $\mathtt{ALPA}$-$\mathtt{finetune}$}}    & \multirow{2}{*}{{75.04}}   & \XSolidBrush    & 50.56 & 82.45   \\  
 &       &\textcolor{red}{\CheckmarkBold} & \cellcolor{gray!20}\textbf{47.53}  &  \cellcolor{gray!20}{84.44}   \\ 
 \midrule 
\multirow{2}{*}{\textbf{CL}}     & \multirow{2}{*}{73.65}   & \XSolidBrush    &60.22  &86.14    \\  
 &       &{\CheckmarkBold} & 60.54  &  85.79   \\   
\multirow{2}{*}{\textbf{+ $\mathtt{ALPA}$-$\mathtt{train}$}}    & \multirow{2}{*}{74.26}   & \XSolidBrush    & 59.02 & 83.74   \\  
 &       &\textcolor{red}{\CheckmarkBold} &\cellcolor{gray!20}55.75   &\cellcolor{gray!20}86.38     \\  
\multirow{2}{*}{\textbf{+ $\mathtt{ALPA}$-$\mathtt{finetune}$}}    & \multirow{2}{*}{\textbf{75.28}}   & \XSolidBrush    & 54.31 & 85.89   \\  
 &       &\textcolor{red}{\CheckmarkBold} & \cellcolor{gray!20}\textbf{50.53}  & \cellcolor{gray!20}\textbf{87.90}    \\   
 \bottomrule
\end{tabular}
\label{ablation}
\end{table}

\begin{figure}[t]
    \centering
    \includegraphics[width=0.49\textwidth]{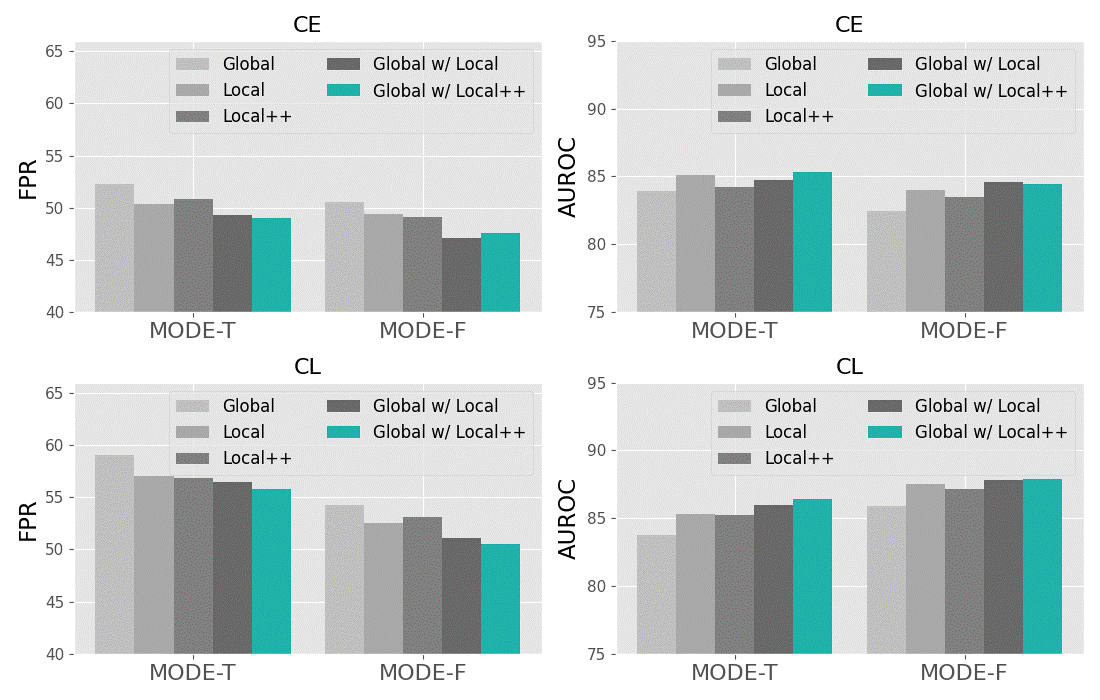}
    \caption{Effect of multi-scale (i.e., global, local and local++) representations on performance -- {CIFAR-100 (ID) with ResNet-34}. 
    The results are the average across five common OOD datasets. For FPR (resp. AUROC), smaller (resp. higher) values are better.}
    \label{region}
\end{figure}

\begin{figure}[t]
    \centering
    \includegraphics[width=0.48\textwidth]{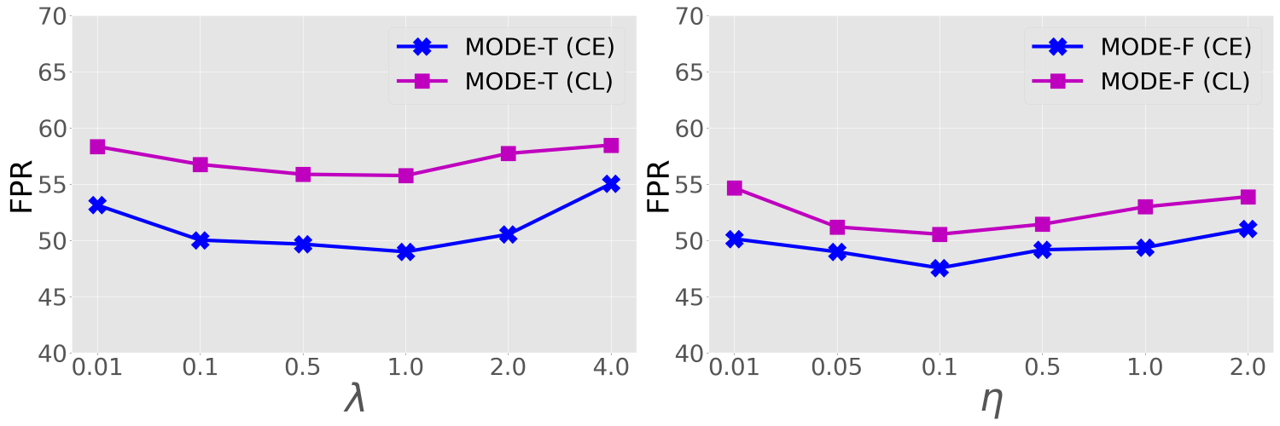}
    \caption{Effectss of the balance weight $\lambda$ for MODE-T and the learning rate $\eta$ for MODE-F on performance -- {CIFAR-100 (ID) with ResNet-34}. 
    The FPR values are the average on the five common OOD datasets, smaller is better.}
    \label{ihyper}
\end{figure}

\subsection{Ablation Studies}
\label{sablation}
In this section, we first conduct ablative analysis to validate the effectiveness of designed components of our MODE in Table \ref{ablation}. 
Then, we analyze the effects of 
\textbf{i}) multi-scale representations, 
\textbf{ii}) balance weight $\lambda$, 
\textbf{iii}) learning rate $\eta$,
and \textbf{iv}) $k$-NN hyperparameter $k$ to deeply investigate our MODE.

\begin{figure*}[t]
    \centering
    \includegraphics[width=1\textwidth]{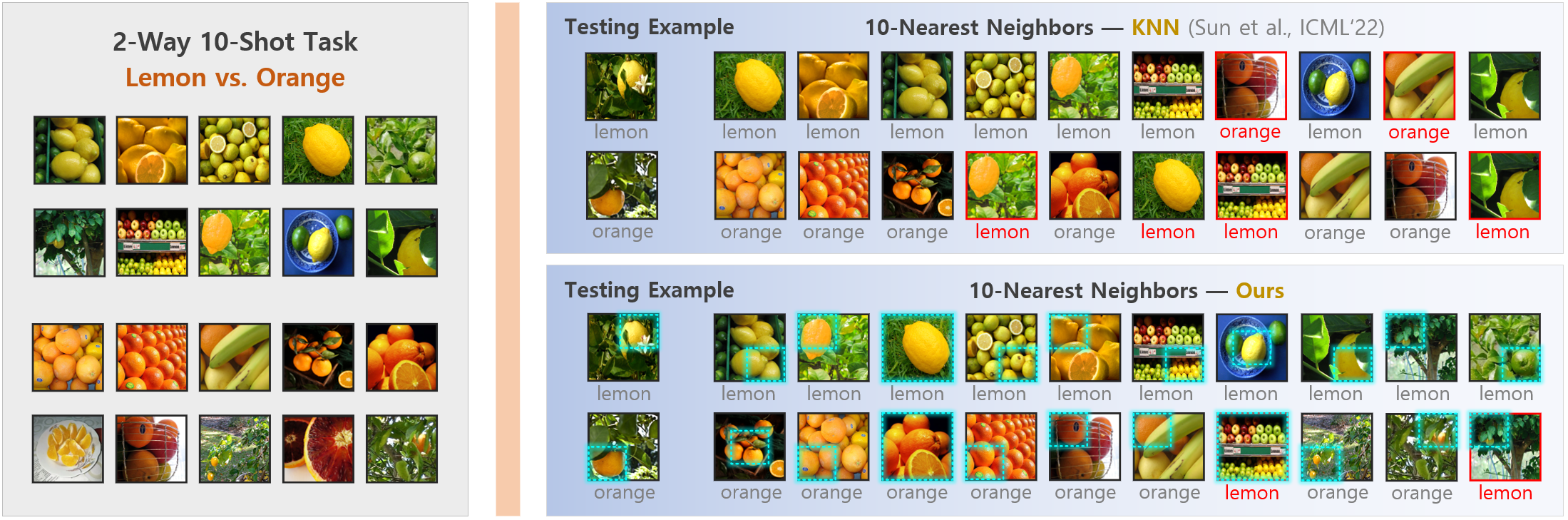}
    \caption{Visualization analysis on $k$-nearest neighbors. For illustrative purposes, we construct a \textit{hard} 2-Way 10-Shot task, which consists of two \textit{near} categories of “\textit{Lemon}” and “\textit{Orange}” from ImageNet-1k \cite{deng2009imagenet}, with 10 examples/class. 
    \textit{Particularly, when “\textit{Lemon}” is treated as ID, “\textit{Orange}” becomes OOD, and vice versa.}
    For each testing example, we search the $k$-nearest neighbors from the 20 seen examples (\textit{left}) in the representation space using KNN \cite{sun2022out} or MODE-F. 
    The {\textcolor{cyan}{cyan}} dotted boxes indicate the image regions corresponding to the most discriminative representations recognized by our method \textcolor{black}{(the results of both the two settings i) ID=“\textit{Lemon}”, OOD=“\textit{Orange}” and ii) ID=“\textit{Orange}”, OOD=“\textit{Lemon}” are shown)}.
    As seen, our method locates the target object regions in the vast majority of cases. Besides, when $4\le k\le10$, totally {\textcolor{red}{6}} and {\textcolor{red}{2}} OOD examples are wrongly detected as ID data for KNN \cite{sun2022out} and our method, respectively.}
    \label{vis}
\end{figure*}

\keypoint{Effectiveness of the Designed Components of MODE.} 
Here, we seek to answer the following two questions:
\textcircled{1} Can our training-time $\mathtt{ALPA}$ encourage locally discriminative representations during ID training?
\textcircled{2} Can our test-time $\mathtt{CSD}$ further boost test-time OOD detection? 
To this end, we conduct experiments on the ResNet-34 feature backbone and use CIFAR-100 as the ID dataset. 
The average results w.r.t. ID ACC, FPR and AUROC on five common OOD benchmarks are reported in Table \ref{ablation}.
We have the following observations.
\underline{\textbf{First}}, from the cells of “ID ACC”, it is obvious that both  $\mathtt{ALPA}$-$\mathtt{train}$ and $\mathtt{ALPA}$-$\mathtt{finetune}$ improve the in-distribution classification performance, which indicates that the designed $\mathtt{ALPA}$ benefits the learning of discriminative representations from ID classes. 
This is in accordance with the historical evidence that local representations (i.e. dense features) inside images can provide richer and more flexible information about the target objects \cite{li2019revisiting,zhang2020deepemd,wang2021dense}. 
More qualitative results are presented in Fig. \ref{vis} and discussed in Section \ref{svis}.
\underline{\textbf{Second}}, from the cells of “$+\mathtt{CSD}$”, we can see that our $\mathtt{CSD}$ significantly improves the test-time OOD detection results with multi-scale representations learned by both $\mathtt{ALPA}$-$\mathtt{train}$ and $\mathtt{ALPA}$-$\mathtt{finetune}$, suggesting the efficacy of our $\mathtt{CSD}$ function.
\underline{\textbf{Third}}, $\mathtt{CSD}$ does not result in significant performance gains on single-scale image representations learned by off-the-shelf cross-entropy (CE) or contrastive learning (CL) losses, which proves the necessity of addressing the scale-discrepancy between ID training and OOD detection, and also confirms the effectiveness of our $\mathtt{ALPA}$ for tackling this problem.
\underline{\textbf{Fourth}}, from the cells of “FPR” and “AUROC”, it can be observed that each of $\mathtt{ALPA}$-$\mathtt{train}$ + $\mathtt{CSD}$ and $\mathtt{ALPA}$-$\mathtt{finetune}$ + $\mathtt{CSD}$ outperforms the baseline methods (i.e., vanilla CE/CL-trained models with KNN-based OOD detection \cite{sun2022out}) by large margins, 
clearly demonstrating the effectiveness of the designed components (i.e. training-time $\mathtt{ALPA}$ and test-time $\mathtt{CSD}$), as well as the flexibility of our proposed MODE framework.

\begin{table}[t]
\centering
\tabcolsep 0.055in
\caption{Inference time of the state-of-the-art KNN \cite{sun2022out} and our \textbf{MODE-F} (the results of \textbf{MODE-T} have similar trends) -- {CIFAR-100 (ID) with ResNet-34}. 
$ ^{\uparrow} $ (resp. $ ^{\downarrow} $) indicates larger (resp. smaller) is better. 
We randomly sampling $\alpha\%$ training data from each ID class for nearest neighbor search.
We conduct the experiment on an \textit{NVIDIA GeForce RTX 3090}.
\\}
\begin{tabular}{cccc}
\toprule
{{\textbf{Method}}}  & {{\textbf{Infer. Time ({ms}/{img})$\downarrow$}}}  & \textbf{FPR$\downarrow$} & \textbf{AUROC$\uparrow$}  \\
\midrule
KNN \cite{sun2022out} & 0.14 & 60.22 &  86.14 \\
\rowcolor{gray!20}{\textbf{MODE-F} ($\alpha=5\%$)\, }  & 0.12  &  54.14 & 86.82  \\ 
{\textbf{MODE-F} ($\alpha=10\%$)}  & 0.25  & 53.18 & 87.15 \\ 
{\textbf{MODE-F} ($\alpha=50\%$)}  & 0.74  & 52.75 & 87.68  \\ 
{\, \textbf{MODE-F} ($\alpha=100\%$)}  & 1.51  &  {50.53}  &  {87.90} \\ 
\bottomrule
\end{tabular}
\label{time}
\end{table}

\begin{figure}[t]
    \centering
    \includegraphics[width=0.48\textwidth]{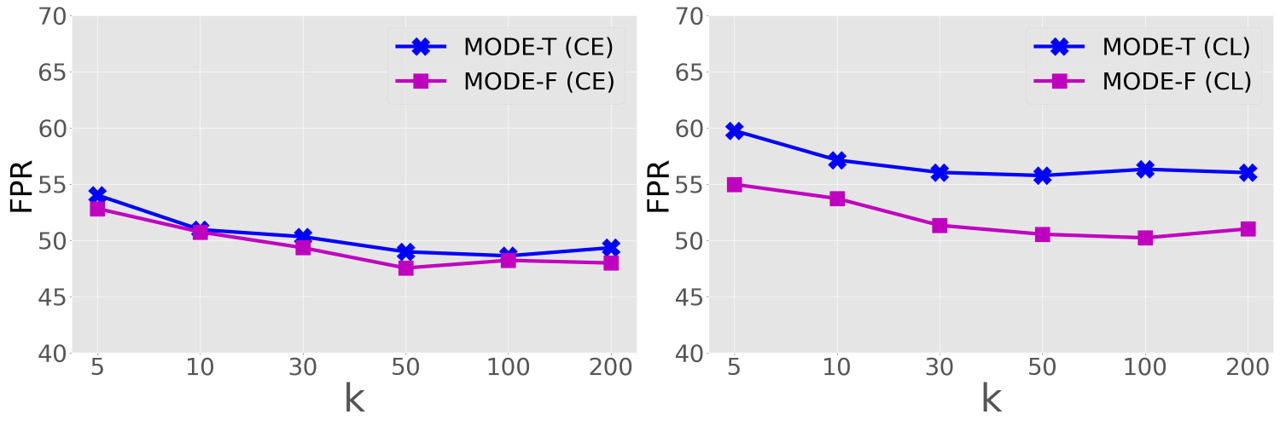}
    \caption{Effect of the KNN hyperparameter $k$ on OOD detection performance -- {CIFAR-100 (ID) with ResNet-34}. The FPR values are the average on five common OOD datasets, smaller is better.}
    \label{imagek}
\end{figure}

\begin{figure}[t]
    \centering
    \includegraphics[width=0.49\textwidth]{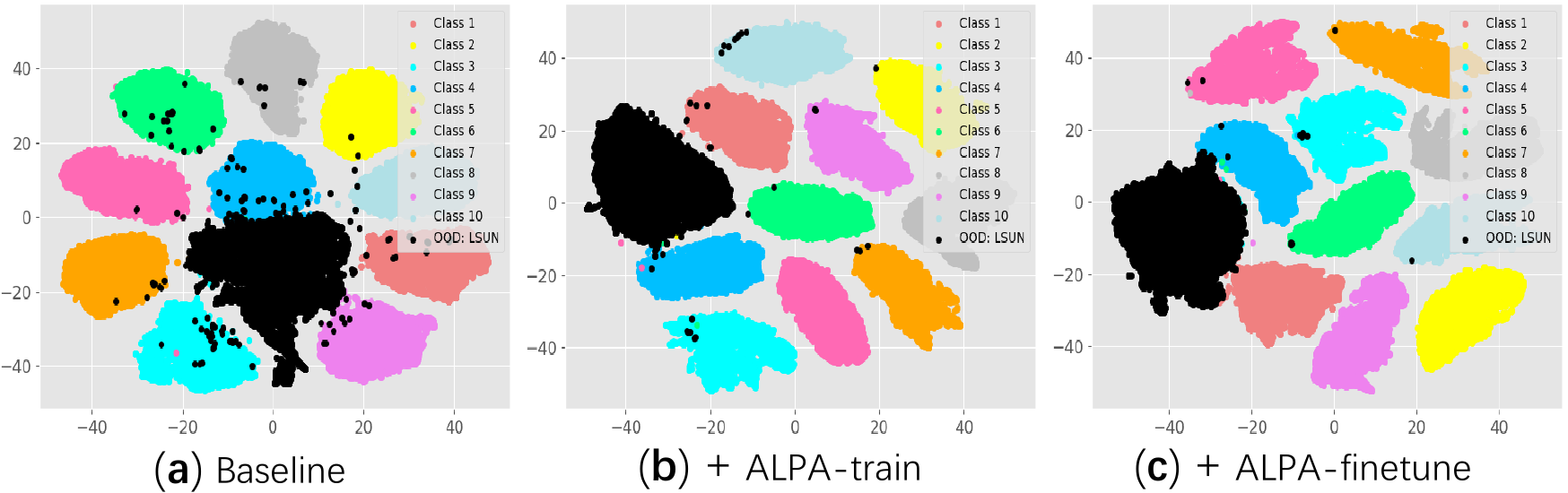}
    \caption{The tSNE visualization of the vanilla CE-trained and our $\mathtt{ALPA}$-enhanced image representations from the penultimate layer of the feature backbone -- CIFAR-10 (ID) with ResNet-18.
    The OOD data LSUN is colored in \textbf{black}, and the ID data CIFAR-10 is denoted by \textbf{non-black} colors.}
    \label{tsne}
\end{figure}

\keypoint{Effect of Multi-scale Representations.} 
During test-time OOD detection, our designed $\mathtt{CSD}$ function explores the most discriminative multi-scale (i.e., both global and local) representations  to distinguish ID/OOD examples more faithfully.
The extracted local representations of each input image $\mathbf{\textbf{x}}$ are concretely the output feature maps before the final global average pooling layer of convolutional networks (or feature backbones), denoted as $\mathbf{\textbf{f}} \in \mathbb{R}^{HW\times E}$.
Therefore, an $M\times M$ image can be mapped into $HW$ local representations (or split image regions), with a corresponding region size $M/H \times M/W$.
That means the larger the number of local representations, the smaller the size of a region. 
As aforementioned in Section \ref{smode} and depicted in Fig. \ref{nag}, we can reduce the number of extracted local representations for every image from $HW$ to $HW/4+1$ (concretely, from $4\times4$ to 5 in our experiments), by performing a {neighbor aggregation} procedure on every four nearest local representations at different positions, and obtain the neighbor aggregated local representations, called local++ representations.
In Fig. \ref{region}, we investigate the effect of those multi-scale (i.e., global, local and local++) representations on the OOD detection performance of our MODE framework.
We have several important observations from the figure. 
\underline{\textbf{First}}, compared with the $\mathtt{ALPA}$-enhanced global representations, the $\mathtt{ALPA}$-enhanced local representations are more beneficial to improve OOD detection performance, which reveals the fact that our $\mathtt{ALPA}$ enables locally discriminative representations that captures richer and more flexible representations for recognizing ID/OOD data.
\underline{\textbf{Second}}, leveraging both global and local representations from images can further boost the results. This is in accordance with our intention that exploiting multi-scale representations from images help maximally benefit OOD detection.
\underline{\textbf{Third}}, the combination of global and local++ representations achieves the best performance in most cases. 
One possible reason is that when an image is divided into a larger number of (i.e. $HW$) local representations, the size of every corresponding region becomes smaller, as a consequence, some of those split regions fail to capture the target objects. 
Therefore, for higher performance and computational efficiency, in our experiments the multi-scale representations for each image specifically include one global representation and five local++ representations.

\keypoint{Effect of the Balance Weight $\lambda$ for MODE-T.} 
During ID training, MODE-T (i.e. $\mathtt{ALPA}$-$\mathtt{train}$ + $\mathtt{CSD}$) employs $\mathtt{ALPA}$-$\mathtt{train}$ to encourage discriminative local representations by regularizing existing ID training loss functions with the devised $\mathcal{L}_{alpa}$, i.e., $\mathcal{L} = \mathcal{L}_{base} + \lambda \mathcal{L}_{alpa}$ in Eq. \ref{lll}, where the hyperparameter $\lambda$ is adopted to balance the contribution of our $\mathcal{L}_{alpa}$.
In this part, we carefully tune $\lambda$ by setting it to the values of \{0.001, 0.1, 0.5, 1.0, 2.0, 4.0\}, and report the average testing results on five common OOD datasets in Fig. \ref{ihyper} (\textit{left}).
As can be observed, our MODE-T is not sensitive to the change of $\lambda$ within a certain range (from 0.1 to 2.0). 
It's worth noting that when $\lambda$ takes the value of 1.0 our MODE-T establishes the best OOD detection performance with both CE-trained and CL-trained baseline models. 
We thus set $\lambda=1.0$ for MODE-T in our experiments.

\keypoint{Effect of the Learning Rate $\eta$ for MODE-F.} 
The most important hyperparameter of our MODE-F (i.e., $\mathtt{ALPA}$-$\mathtt{finetune}$ + $\mathtt{CSD}$) is the initial learning rate $\eta$ for finetuning pretrained models learned by different losses, using the developed $\mathtt{ALPA}$-$\mathtt{finetune}$.
To investigate the effect of $\eta$ on the performance MODE-F, we carefully tune $\eta$ by setting it to different values of \{0.001, 0.05, 0.1, 0.5, 1.0, 2.0\}. 
We report the average results on five common OOD datasets in Fig. \ref{ihyper} (\textit{right}).
As seen, our MODE-F achieves remarkable and stable performance when $\eta$ takes the values within a certain range (from 0.05 to 0.5). 
In particular, when $\eta=0.1$ our MODE-F achieves the best results with both CE-trained and CL-trained baselines, we therefore set $\eta=0.1$ for MODE-F in our experiments.

\keypoint{Effect of the $k$-NN Hyperparameter $k$.} 
Both KNN \cite{sun2022out} and our MODE ($\mathtt{CSD}$, more concretely) need to adjust the $k$-NN hyperparameter $k$.
In Fig. \ref{imagek}, we analyze the effect of $k$ on the OOD detection performance of our MODE.
Specifically, we carefully tune $k$ by setting it to the values of \{5, 10, 30, 50, 100, 200\}, and report the average results on five common OOD datasets.
As can be observed from the figure, the OOD detection performance gradually improves with the increase of $k$ before $k$ reaches 50. 
This trend is also consistent with the ablation results of $k$ in KNN \cite{sun2022out} under the same setting. 
Additionally, we also observe that the OOD detection results of both MODE-T and MODE-F remain similar when $k$ takes the values of 50, 100 and 200. Hence, in our experiments, we set $k=50$ as in KNN \cite{sun2022out}.

\subsection{Computational Cost} 
\label{stime}
It is important to study the computational cost of our proposed MODE for practical purposes.
In this part, we quantitatively investigate the test-time OOD detection computational cost of our MODE (concretely, brought by the test-time $\mathtt{CSD}$), according to the per-image inference time (in \textit{milliseconds}). In particular, we randomly sample $\alpha\%$ training data from each class of the ID dataset (i.e., CIFAR-100 with 50,000 training examples) for $k$-nearest neighbor search on testing OOD data. 
We report the per-image inference time of MODE-F (while the results of {MODE-T} have similar trends) at different values of $\alpha\%$ in Table \ref{time}, where we conduct the experiment on an \textit{NVIDIA GeForce RTX 3090}.
It should be noted that when $\alpha$ takes the values of \{5, 10, 50, 100\}, we set the $k$-NN hyperparameter $k$ to \{10, 20, 30, 50\} for our MODE-F, respectively. 
We highlight four important observations in the table.
\underline{\textbf{First}}, when $\alpha=100\%$, the per-image inference time of our method is 1.51 \textit{milliseconds}, a result that may be acceptable in many real-world (offline or online) applications.
\underline{\textbf{Second}}, as expected, the inference time cost of our method gradually decreases as $\alpha$ decreases.
\underline{\textbf{Third}}, the plunge in $\alpha$ does not severely degrade the OOD detection performance of our method.
\underline{\textbf{Fourth}}, when spending a comparable time consumption as the state-of-the-art KNN \cite{sun2022out}, our method still outperforms KNN by a large margin (i.e., \textbf{6.08\%} in FPR). 
All the above results suggest that our proposed MODE enjoys good practicability and scalability.

 \subsection{Visualization Analysis}
\label{svis}
So far, we have quantitatively demonstrated the effectiveness and flexibility of our developed MODE framework for OOD detection. In this part, we present some visualization results to qualitatively investigate our MODE.

\keypoint{Visualization with tSNE \cite{van2008visualizing}.}
In Fig. \ref{tsne}, we present the tSNE visualization of the vanilla CE-trained and our $\mathtt{ALPA}$-enhanced global representations (extracted from the penultimate layer of the feature backbone ResNet-18) of the ID dataset CIFAR-10 and the OOD data LSUN (OOD)  
-- the results of vanilla CL-trained representations have similar trends.
As can be observed from the figure, compared with the vanilla CE-trained global representations, the global representations learned by each of our $\mathtt{ALPA}$-$\mathtt{train}$ or $\mathtt{ALPA}$-$\mathtt{finetune}$ exhibit better ID-OOD separability.
We also see that although the baseline improves the compactness of each ID class, there is a significant overlap between these ID classes and the OOD data.
Generally speaking, in combination with the quantitative ID classification and OOD detection results in Table \ref{ablation}, it is apparent that our designed $\mathtt{ALPA}$ not merely encourage locally discriminative representations during ID training, but also drive the extracted image representations of different ID/OOD classes to be more compact for benefiting both OOD detection as well as multi-class classification tasks.

\keypoint{Visualization of $k$-Nearest Neighbors.}
In Fig. \ref{vis}, we further demonstrate the effectiveness of our MODE by qualitatively comparing its searched $k$-nearest neighbors with that of KNN \cite{sun2022out} on testing examples. 
In this experiment, leveraging the \{“\textit{Lemon}” vs. “\textit{Orange}”\} task for visualization analysis is inspired by the fact that \textit{hard} OOD detection tasks composed with near ID-OOD classes/examples are the major challenge for existing machine learning systems, as revealed in \cite{fort2021exploring,sastry2020detecting}.
In this task, when  we treat “\textit{Lemon}” as ID data, “\textit{Orange}” becomes OOD, and \textit{vice versa}.
As can be observed from the figure, our method successfully identifies the most discriminative multi-scale representations corresponding to the target objects (or object regions) in the vast majority of cases. 
What is noteworthy is that local representations/regions play a key role in successfully recognizing those hard examples with cluttered backgrounds.
Moreover, when $4\le k\le10$, totally {{6}} and {{2}} OOD examples are wrongly detected as ID data for KNN \cite{sun2022out} and our method, respectively.
All in all, benefiting from highlighting richer and more transferable representations during ID training (by $\mathtt{ALPA}$), and taking advantage of the most discriminative multi-scale representations for test-time OOD detection (by $\mathtt{CSD}$), our proposed MODE shows remarkable performance on distinguishing ID/OOD data. 

\section{Conclusion}
For the first time, this work proposes MODE to leverage multi-scale representations inside images for OOD detection.
Concretely, we first observe that due to the scale-discrepancy between the ID training and OOD detection  processes, existing models pretrained by off-the-shelf cross-entropy or contrastive losses are unable to capture usable local representations for MODE.
To address this issue, we propose \textbf{$\mathtt{ALPA}$}, which enables locally discriminative representations by aligning and highlighting the local object regions of pairwise examples during ID training. 
During test-time OOD detection, we devise a \textbf{$\mathtt{CSD}$} function on the most discriminative multi-scale representations to distinguish ID/OOD examples more faithfully.
Our MODE framework is orthogonal to ID training losses and models pretrained with different fashions. 
Extensive experimental results demonstrate the effectiveness and flexibility of our MODE on a wide range of baseline methods applied to various network structures. 
We hope this work can bring new inspiration to OOD detection as well as other related fields.
To facilitate future research, we have made our code publicly available at: \url{{https://github.com/JimZAI/MODE-OOD}}.

\bibliographystyle{IEEEtran}
\bibliography{tmm}
\end{document}